\documentclass{cls/IEEEojcsys}

\usepackage[colorlinks,urlcolor=blue,linkcolor=blue,citecolor=blue]{hyperref}
\usepackage{color,array}

\usepackage{graphicx}

\receiveddate{}
\reviseddate{}
\accepteddate{}
\publisheddate{}
\currentdate{}
\editor{}
\makeatletter
\let\@revised\@empty
\makeatother

\usepackage{amsthm}

\usepackage[most]{tcolorbox}
\newtheorem{theorem}{Theorem}

\newtheorem{definition}{Definition}
\newtheorem{remark}{Remark}
\newtheorem{problem}{Problem}

\usepackage{hyperref} 
\usepackage{amsmath,amssymb,amsfonts,mathtools}
\usepackage{bbm}

\usepackage[ruled,vlined,linesnumbered]{algorithm2e} % algorithm

\usepackage{xcolor}

\def\BibTeX{{\rm B\kern-.05em{\sc i\kern-.025em b}\kern-.08em
    T\kern-.1667em\lower.7ex\hbox{E}\kern-.125emX}}

\usepackage{lipsum}
\makeatletter
\newcommand{\algorithmfootnote}[2][\footnotesize]{%
  \let\old@algocf@finish\@algocf@finish% Store algorithm finish macro
  \def\@algocf@finish{\old@algocf@finish% Update finish macro to insert "footnote"
    \leavevmode\rlap{\begin{minipage}{\linewidth}
    #1#2
    \end{minipage}}%
  }%
}
\makeatother

\makeatletter
\renewcommand{\p@subsection}{\thesection-}
\makeatother

\newcommand{\sndbest}[1]{{\color{black!50}\textit{\textbf{#1}}}}

\newcommand{\calA}{{\cal A}}
\newcommand{\calB}{{\cal B}}

\newcommand{\calD}{{\cal D}}

\newcommand{\calF}{{\cal F}}

\newcommand{\calL}{{\cal L}}

\newcommand{\calO}{{\cal O}}
\newcommand{\calP}{{\cal P}}

\newcommand{\calS}{{\cal S}}
\newcommand{\calT}{{\cal T}}

\newcommand{\calX}{{\cal X}}

\newcommand{\bfa}{\mathbf{a}}
\newcommand{\bfb}{\mathbf{b}}

\newcommand{\bfe}{\mathbf{e}}
\newcommand{\bff}{\mathbf{f}}
\newcommand{\bfg}{\mathbf{g}}
\newcommand{\bfh}{\mathbf{h}}

\newcommand{\bfp}{\mathbf{p}}

\newcommand{\bfs}{\mathbf{s}}

\newcommand{\bfu}{\mathbf{u}}
\newcommand{\bfv}{\mathbf{v}}

\newcommand{\bfx}{\mathbf{x}}
\newcommand{\bfy}{\mathbf{y}}
\newcommand{\bfz}{\mathbf{z}}

\newcommand{\bfdelta}{\boldsymbol{\delta}}

\newcommand{\bfkappa}{\boldsymbol{\kappa}}

\newcommand{\bftau}{\boldsymbol{\tau}}

\newcommand{\bfpsi}{\boldsymbol{\psi}}

\newcommand{\bfxi}{\boldsymbol{\xi}}

\newcommand{\bfT}{\mathbf{T}}

\newcommand{\bbR}{\mathbb{R}}

\newcommand{\red}[1]{\textcolor{red}{#1}}

\newcommand{\blue}[1]{\textcolor{blue}{#1}}
\newcommand{\cyan}[1]{\textcolor{cyan}{#1}}

\newcommand{\orange}[1]{\textcolor{orange}{#1}}

\newcommand{\Kurkova}{K\.urkov\`a}

\usepackage{multirow}
\newcommand{\defeq}{\vcentcolon=}
\newcommand{\MLP}{\text{MLP}}
\newcommand{\spline}{\text{sp}}
\newcommand{\ReLU}{\text{ReLU}}
\newcommand{\teq}{\coloneqq}
\DeclareMathOperator*{\argmin}{arg\,min}

\usepackage{hyperref}
\usepackage{xcolor}
\usepackage[enable]{easy-todo} % disbale to disable

\usepackage{colortbl}
\usepackage{booktabs}

\definecolor{darkgreen}{HTML}{008000}  % define once, reuse
\definecolor{darkbrown}{HTML}{b36318}
\definecolor{lightpurple}{HTML}{b31894}

\definecolor{lightblue}{rgb}{0.83,0.85,1.0}
\definecolor{light-gray}{gray}{0.95}

\definecolor{rowgray}{gray}{0.95}
\definecolor{headerblue}{rgb}{0.784, 0.784, 1.0} % 200/255 = 0.784
\definecolor{lightgray}{gray}{0.95}

\newcolumntype{T}{>{\columncolor{lightblue}}c}
\newcolumntype{V}{>{\columncolor{light-gray}}c}

\newcommand{\TwoRow}[2]{
    #1 \\ 
    \rowcolor{lightgray}
    #2 \\
}
\newcommand{\ARow}[4]{
\cellLeft{#1} & #2 & & \cellLeft{#3} & #4 &
}

\usepackage{acro}
\DeclareAcronym{NN}{short = NN, long = neural network}
\DeclareAcronym{KAN}{short = KAN, long = Kolmogorov-Arnold network}
\DeclareAcronym{KAT}{short = KAT, long = Kolmogorov-Arnold theorem}
\DeclareAcronym{KKAN}{short = KKAN, long = \Kurkova{} Kolmogorov-Arnold network}
\DeclareAcronym{GP}{short = GP, short-plural-form = GPs, long = Gaussian process, long-plural-form = Gaussian processes}
\DeclareAcronym{MLP}{short = MLP, long = multi-layer perceptron, short-indefinite = a, long-indefinite = a}
\DeclareAcronym{RBF}{short = RBF, long = radial basis function}
\DeclareAcronym{SNR-MLP}{short = SNR-MLP, long = spectral-normalized ReLU activated MLP}
\DeclareAcronym{DAREK}{short = DAREK, long = distance-aware error for Kolmogorov networks}
\DeclareAcronym{K-DAREK}{short = K-DAREK, long = distance-aware error for \Kurkova{} Kolmogorov networks}
\DeclareAcronym{DKL}{short = DKL, long = deep kernel learning}
\DeclareAcronym{SNGP}{short = SNGP, long = spectral-normalized neural Gaussian process}
\DeclareAcronym{DUE}{short = DUE, long = deterministic uncertainty estimation}
\DeclareAcronym{DUQ}{short = DUQ, long = deterministic uncertainty quantification}
\DeclareAcronym{DDU}{short = DDU, long = deep deterministic uncertainty}
\DeclareAcronym{ReLU}{short = ReLU, long = rectified linear unit}
\DeclareAcronym{MSE}{short = MSE, long = mean squared error}
\DeclareAcronym{MPC}{short = MPC, long = model predictive control}
\DeclareAcronym{CBF}{short = CBF, long = control barrier function, long-plural-form = control barrier functions}
\DeclareAcronym{SNN}{short = SNN, long = spline neural network, long-plural-form = spline neural networks}
\DeclareAcronym{BNN}{short = BNN, long = Bayesian neural network, long-plural-form = Bayesian neural networks}
\begin{document}
%%%%%%%%%% OJCSys %%%%%%%%%%%
\sptitle{Article Category}

\title{Worst-Case Distance-Aware Error Bounds for Neural Networks}

\author{Masoud Ataei\affilmark{1}}
\author{Vikas Dhiman\affilmark{1} (Member, IEEE)}
\author{Mohammad Javad Khojasteh\affilmark{2}  (Member, IEEE)}

\affil{Electrical and Computer Engineering, University of Maine, Orono, ME 04469 USA} 
\affil{Electrical and Microelectronic Engineering, Rochester Institute of Technology, Rochester, NY 14623 USA} 

\corresp{CORRESPONDING AUTHOR: M. Ataei (e-mail: \href{masoud.ataei@maine.edu}{masoud.ataei@maine.edu})}
\authornote{This work is supported by the National Science Foundation under Grant No. 2218063 and the Gleason Endowment at RIT.}

%%%%%%%%%%%%%%%%

\begin{abstract}
Safety-critical applications of machine learning require uncertainty estimates that support reliable worst-case analysis.
\Acp{NN} provide expressive function approximation, while \acp{GP} offer principled probabilistic uncertainty, but both face limitations in this setting: most modern neural architectures lack tractable worst-case error bounds, and Gaussian processes become computationally expensive at scale.
For uncertainty to be interpretable, a central requirement is distance-awareness: that is, uncertainty increases with the distance between a test input and the nearest relevant training data.
We present a general framework for worst-case distance-aware error bounds for \acp{NN} that combine dense layers with spline-based components. 
Our approach establishes error bounds that are both distance-aware, reflecting proximity of a test point to its nearest training data, and worst-case, providing deterministic guarantees under {known} Lipschitz constraints rather than probabilistic assumptions.
Our algorithm, K-DAREK (Distance-Aware Error for \Kurkova{}-Kolmogorov Networks), provides efficient and interpretable uncertainty quantification for \acp{NN}.
K-DAREK is about four times faster and ten times more computationally efficient than an ensemble of \acsp{KAN},
8.6 times more scalable than \ac{GP}, 
eliminates up to 8.2\% error-bound violation rate observed in DAREK, and reduces the average collision rate from 1.8\% to 1.1\% in the multi-agent safe control experiment.
On high-dimensional 
real-world regression tasks (e.g., Real Estate Valuation), K-DAREK preserves distance-aware {error bounds}
and achieves zero coverage violations, 
addressing the overgeneralization and inducing-point-coverage limitations exhibited by SNGP and DUE, respectively. 
\end{abstract}

\maketitle

\acresetall % Reprints full abbreviations

%--------------------- Table of Symbols ------------------------%
\begin{table*}[ht]
\caption{Notation and definitions of important quantities. 
} 
\label{tbl:symbol_notation}
\setlength{\tabcolsep}{1pt} % default value: 6pt
\renewcommand{\arraystretch}{1.3}
\providecommand {\cellLeft} [1] {\hspace{0.2cm}#1}
\centering
\begin{tabular}{  m{2.5cm}  m{6.0cm} m {0cm} |  m{2.5cm} m{6.0cm} m{0cm}}%\\[0.4ex]
\hline %\vspace{0.25ex}
\rowcolor{headerblue} 
{\textbf{Notation}} & \textbf{Definition} & &
{\textbf{Notation}} & \textbf{Definition}  & \\[0.25ex]
\hline

%---------------------  Two Rows  ---------------------%
\TwoRow{
    \ARow{$f$, $\hat{f}$}{Target function and trained approximator}
         {$\hat{h}_{\MLP}$, $\hat{h}_{\spline}$}{MLP block and spline block of K-DAREK}}
{   \ARow{$d$, $d_u(.)$}{Input dimension $\bfx \in \bbR^d$, distance function candidate for uncertainty function $u$}
         {$q$}{Intermediate feature dimension}}
%---------------------  Two Rows  ---------------------%
\TwoRow{\ARow{$\calD$}{Training dataset, $\{(\bfx_i, y_i)\}_{i=1}^n$}
             {$L$}{Number of layers in each SNR-MLP}
             }
       {\ARow{$\calX$, $\calX_\calD$}{Input space ($\calX \subset \bbR^d$) and input portion of $\calD$}
             {$\bftau$, $k$}{Set of knots and order of spline}}
%---------------------  Two Rows  ---------------------%
\TwoRow{\ARow{$\bfx$, $\bfx^*$}{Input vector and test input vector}
       {$\bftau^*$ }{Nearest training input (knot) to test point $\bfx^*$}
             }
       {\ARow{$\calL_f$, $\calL_{\spline}$}{Lipschitz constant of true function $f$ and spline block}
       {$\calL_{\MLP}$, $\calL_{\MLP_i}$}{Lipschitz constants of MLP block and the $i$-th SNR-MLP}}
             
%---------------------  Two Rows  ---------------------%
\TwoRow{\ARow{$u_{\MLP}$, $u_{\spline}$, $u_f$}{Error bounds on MLP block, spline block, and target function $f$}
             {$\calT$, $K$}{Input knot matrix ($\calT \in \bbR^{m_k \times d}$) and feature-space knot matrix ($K \in \bbR^{m_k \times q}$).}}
       {\ARow{$\bfxi$, $\bfxi_i$}{Aggregated latent feature vector before spline mapping and its $i$-th component}
             {$m_k$}{Number of knots per spline}}
\hline
\end{tabular}
\end{table*}

\section{INTRODUCTION}
Deploying \acp{NN} in safety-critical applications requires provable error bounds, which most modern architectures lack~\cite{ovadia2019can}.
\Ac{NN} predictions do not generally provide guarantees on the worst-case deviation from the true underlying function, making it difficult to reason about safe decisions or operations. 
A principled worst-case analysis for \acp{NN} at any test point would quantify the maximum possible deviation between the network's prediction and the true function.
We present a general framework for a worst-case distance-aware error bound for \acp{NN} that combine dense layers with spline-based components.
Splines provide smooth, piecewise polynomial mappings that ensure local control and continuous derivatives, making them well-suited for modeling complex functions, particularly in control engineering and signal processing~\cite{wang1995bootstrap,egerstedt2009control,unser2002splines,fey2018splinecnn,wang2011asymptotics,balestriero2018spline,wahba1975smoothing}. 
Spline-based \acp{NN}~\cite{igelnik2003kolmogorov,bohra2020learning,aziznejad2020deep,wang2026physics} can leverage these properties to produce smoother function approximations than conventional \acp{MLP}.
Furthermore, spline-based \acp{NN} offer improved interpretability and thus greater reliability~\cite{liu2025kan}. However, producing robust and trustworthy uncertainty estimates remains difficult.
Two predominant strategies for uncertainty estimation are probabilistic methods and worst-case formulations (e.g., adversarial or robust).

Probabilistic methods are generally accurate in data-rich domains, particularly when the goal is to model stochastic behavior.
Common methods for probabilistic uncertainty estimation include \acp{GP}~\cite{rasmussen2006gaussian} and Bayesian neural network approaches such as Monte Carlo dropout~\cite{gal2016dropout}, variational inference~\cite{blundell2015weight}, and ensembles~\cite{lakshminarayanan2017simple}.
Interpreting the output of these methods often requires calibration, which presents challenges, including that their distributions are typically unbounded, unscaled, and valid only for the observed portion of the input space~\cite{guo2017calibration, jiang2023knowledge,cheng2021limits,ataei2024dadee}.
Additionally, \acp{GP}, which are popular in uncertainty estimation, are computationally expensive and also rely on kernel hyperparameters~\cite{rasmussen2006gaussian}. 
On the other hand, worst-case or robust approaches, which are particularly relevant for safety-critical applications, offer certainty guarantees under specified constraints or perturbations~\cite{jaulin2001interval,ataei2025darek,lopez2023towards,bertsimas2021robust,bertsimas2019robust}. 
These models are often easier to understand, debug, and implement.
While they can sometimes be conservative and face challenges in scaling to high-dimensional or complex distributions, they remain a practical and reliable choice when system constraints are known as hard constraints instead of prior distributions~\cite{ataei2025darek,long2022control}. In control theory, worst-case analysis can be particularly appealing because many problems are formulated in terms of bounded errors~\cite{nair2013nonstochastic,prajna2007framework}.

{An interpretable uncertainty estimator is expected to demonstrate distance-awareness, that is, to have higher predictive confidence in regions near the training data, while exhibiting increased uncertainty as the input deviates from the observed training data~\cite{lakshminarayanan2017simple,liu2020simple, ataei2025darek, mukhoti2023deep, van2020uncertainty}.} 
Uncertainty estimation in nonparametric models---such as \acp{GP}, splines, k-nearest neighbors, kernel density estimation~\cite{bishop2006pattern}---can be easily made distance-aware because they depend on stored training data, whereas typical parametric models---such as ensembles~\cite{lakshminarayanan2017simple} and support vector machines---do not estimate distance-aware uncertainty.

More recently, \ac{KAN}~\cite{liu2025kan,vaca2024kolmogorov,patra2025physics,sen2026physics,li2025kolmogorov,howard2026sindy} introduced a novel approach that combines nonparametric models (splines) with parametric models~\cite{ataei2025darek}. 
\ac{KAN}, inspired by the \ac{KAT}~\cite{schmidt2021kolmogorov,montanelli2020error}, was proposed to model complex functions with a deep neural network using spline bases as activation functions. 
Building on this success, \acp{KKAN}~\cite{toscano2025kkans} reduced the complexity of the \ac{KAN} architecture by replacing its arbitrary-depth spline network structure with a simple two-block architecture, combining an \ac{MLP}-based inner block with flexible basis functions as the outer block. This results in faster, more stable training and improved function approximation. 

However, neither \ac{KAN} nor \ac{KKAN} provides distance-aware uncertainty estimates, even though both rely heavily on nonparametric components in their architectures.
In prior work~\cite{ataei2025darek}, we resolved this issue for \acp{KAN}; here, we further generalize the approach by equipping \acp{KKAN} with distance-aware uncertainty.
Our modified \ac{KKAN} architecture consists of two blocks: an \ac{MLP} block and a spline block (see Fig.~\ref{fig:K-DAREK}).
In the \ac{MLP} block, each input dimension is transformed through a spectrally normalized \ac{MLP} layer~\cite{miyato2018spectral, liu2020simple} to produce higher-dimensional representations.
The resulting representations are summed dimension-wise to form a unified
feature vector. Finally, the spline block maps this feature vector to the output via a linear combination of spline functions.
This unique structure enables learning the \ac{MLP} block with a desired Lipschitz constant. We then apply the spline error analysis from our previous work~\cite{ataei2025darek} to the spline block to compute the error bound for the joint \ac{KKAN} architecture.

%------------------- Table for Acronyms -------------------%
\begin{table}
    \centering    
    \caption{Expansions of important acronyms.
    }
    \label{tbl:acronyms}
    \rowcolors{2}{rowgray}{white}  
    \resizebox{\columnwidth}{!}{%
    \begin{tabular}{ll} 
        \arrayrulecolor{headerblue}\specialrule{\heavyrulewidth}{0pt}{0pt}
        \rowcolor{headerblue}             
            \textbf{Acronym} & \textbf{Expansion}\\         
        \midrule
            \acrodonotuse\acs{NN}      & \acrodonotuse\Acl{NN} \\
            \acrodonotuse\acs{SNN}     & \acrodonotuse\Acl{SNN} \\
            \acrodonotuse\acs{KAT}     & \acrodonotuse\Acl{KAT} \\
            \acrodonotuse\acs{KAN}     & \acrodonotuse\Acl{KAN} \\
            \acrodonotuse\acs{KKAN}    & \acrodonotuse\acl{KKAN} \\
            \acrodonotuse\acs{ReLU}    & \acrodonotuse\Acl{ReLU} \\            
            \acrodonotuse\acs{MLP}     & \acrodonotuse\Acl{MLP} \\           
            MLP-Spline    & A network that uses MLPs and Splines in its architecture\\
            \acrodonotuse\acs{SNR-MLP} & \acrodonotuse\Acl{SNR-MLP} \\
            \acrodonotuse\acs{RBF}     & \acrodonotuse\Acl{RBF} \\
            
            \acrodonotuse\acs{GP}      & \acrodonotuse\Acl{GP} \\
            \acrodonotuse\acs{DKL}     & \acrodonotuse\Acl{DKL}  \\
            \acrodonotuse\acs{DUE}     & \acrodonotuse\Acl{DUE} \\
            \acrodonotuse\acs{SNGP}    & \acrodonotuse\Acl{SNGP} \\
            \acrodonotuse\acs{DAREK}   & \acrodonotuse\Acl{DAREK} \\
            \acrodonotuse\acs{K-DAREK} & \acrodonotuse\acl{K-DAREK} \\
            
            \acrodonotuse\acs{CBF}     & \acrodonotuse\Acl{CBF}  \\
            \acrodonotuse\acs{MPC}     & \acrodonotuse\Acl{MPC}  \\
            \acrodonotuse\acs{MSE}     & \acrodonotuse\Acl{MSE}  \\
            
        \bottomrule
    \end{tabular}
    }
\end{table}

The term ``spline'' originates from the flexible wooden strips once used by shipbuilders, which were shaped by anchoring them to fixed control points (or knots). In spline error analysis, this idea translates into constraining spline knots to a subset of the training data and estimating the approximation error at a test point as a function of its distance to the nearest knots. Within this framework, our prior work~\cite{ataei2025darek}, \Ac{DAREK}, offers a structured, bottom-up framework for uncertainty quantification, allowing error diagnosis at the unit level of a neural network. However, it lacks a principled mechanism for Lipschitz division and error sharing; therefore, its performance is sensitive to heuristic design choices. Moreover, the composition of spline functions in \ac{DAREK}, and in \acp{SNN} more broadly, can yield highly nonlinear behavior, often resulting in non-smooth approximations (which may be mitigated through regularization~\cite{liu2025kan}).
The hybrid architecture of~\ac{K-DAREK}, which combines \ac{MLP} and spline components, produces smoother function approximations by using simpler nonlinear transformations. Also, incorporating scaled spectral normalization introduces explicit Lipschitz control and confines the feature range of each \ac{MLP} block.
Together, these enhancements reduce the challenges of Lipschitz sharing and improve the stability and reliability of uncertainty estimates.

In summary, our goal is to efficiently estimate distance-aware worst-case error bounds for \acp{NN}. 
This is made possible by integrating splines into deep \acp{NN}. 
We propose a general framework for worst-case error bounds for \ac{MLP}-Spline models, with \ac{K-DAREK} as an instance of these models. 
The contributions of this work are as follows:
1) We propose a novel general framework for deriving worst-case error bounds that leverages the expressive power of \acp{MLP} together with spline-based structures to provide interpretable, distance-aware error bound estimates{, under the assumption of known Lipschitz constraints.}
2) We introduce \ac{K-DAREK} as an instance of our framework, and compare \ac{K-DAREK} against \ac{SNGP}~\cite{liu2020simple}, \ac{DUE}~\cite{van2021feature}, \acp{GP}, ensembles, and our previous work~\cite{ataei2025darek}.
{We evaluate K-DAREK on} synthetic datasets, real-world regression datasets (e.g., Real Estate Valuation), 
% for large datasets, 
{as well as in regions with missing data across multiple datasets}.
{We further test on} a high-dimensional face bounding-box prediction and a safe multi-agent control task. 
{We also conduct ablation studies to isolate the contributions of the MLP block, spectral normalization, and the knot-selection strategy, and a sensitivity analysis quantifying how the violation rate degrades when the Lipschitz constant is underestimated.}
We find that \ac{K-DAREK} is about four times \textbf{faster} and ten times more computationally \textbf{efficient} than ensembles,
8.6 times more \textbf{scalable} than \ac{GP},
{eliminates up to an 8.2\% error-bound violation rate observed in DAREK~\cite{ataei2025darek}, and reduces the average collision rate from 1.8\% to 1.1\% in the multi-agent safe control experiment.}
Furthermore, {unlike SNGP,} K-DAREK does not suffer from ``feature collapse'' (overgeneralization)~\cite{van2021feature} on large datasets, effectively captures uncertainty in regions with missing data, and remains distance-aware in high dimensions. 

A preliminary version of this work appeared in~\cite{ataei2025kdarek}. The current paper substantially extends that work in the following ways: 1) we formalize the problem of worst-case distance-aware error bound of general \acp{NN} rigorously and exemplify it in the specific case of \ac{KKAN} model; 2) we add a comprehensive related work section, a structured overview of the framework, and an computational-complexity analysis; 3) we significantly expand the experimental evaluation to cover feature collapse, missing-data regions {across multiple function constructions}, high-dimensional face bounding-box prediction (up to 200 dimensions), a fixed budget comparison, real-world regression datasets, {an ablation study isolating the contribution of each architectural component, a sensitivity analysis of the assumed Lipschitz constant,} and an extensive safe control experiment; 4) we release a software package for reproducing these experiments\footnote{
\href{https://github.com/Masoud-Ataei/KDAREK} {https://github.com/Masoud-Ataei/KDAREK}}.

Important notations and definitions used in the
paper are summarized in Table~\ref{tbl:symbol_notation}.
Acronym expansions are provided in Table~\ref{tbl:acronyms}.

\section{RELATED WORKS}
\label{sec:relatedwork}
Uncertainty estimation in \acp{NN} is a well-investigated research problem. However, most of these models are probabilistic and not distance-aware. For a comprehensive survey, we refer the interested reader to~\cite{abdar2021review,gawlikowski2023survey}.
In this section, we discuss closely related work and provide a historical perspective.

\noindent{\textit{}}\underline{Spline-Based Neural Networks:}
\Ac{KAT} states that any multivariate continuous function can be represented as a finite decomposition of univariate continuous functions and addition~\cite{schmidt2021kolmogorov}. The theorem does not hold if the continuity assumption is replaced with the continuously differentiable assumption~\cite{morris2021hilbert}, and there has been a long debate about whether the \ac{KAT} is relevant to the design of \acp{NN}. 
This controversy is exemplified by two influential papers: Girosi and Poggio argue that \ac{KAT} lacks practical implications for modern network architectures, as it does not account for the smoothness and generalization properties that are critical in practical learning systems~\cite{girosi1989representation}.
Whereas \Kurkova{} contends that the theorem provides foundational insight into the representational capabilities of \acp{NN}~\cite{kuurkova1991kolmogorov,kuurkova1992kolmogorov,schmidt2021kolmogorov}. 
Z. Liu et al.~\cite{liu2025kan} address this issue by focusing on \ac{SNN} with many layers. \acp{KAN}~\cite{liu2025kan} replace traditional fixed activation functions with learnable, spline-parametrized functions on edges, enhancing interpretability and adaptability. 
\Kurkova{} introduced a different remedy for this over three decades ago~\cite{kuurkova1991kolmogorov}. 
She suggested that the component function in the decomposition could be effectively approximated using an \ac{MLP} with small error, rather than \ac{KAT}, where the aim was to achieve zero error.
While \ac{KAN} and \ac{KKAN} improve interpretability, neither inherently provides distance-aware error bound estimates and worst-case error guarantees; our previous work addressed this for \ac{KAN}~\cite{ataei2025darek}, and this paper extends it to \ac{KKAN}.

\noindent{\textit{}}\underline{Deep kernel learning:}
There are similarities between Deep Kernel Learning (DKL) and our design, which is based on \ac{KKAN}.
Wilson et al.~\cite{wilson2016deep} introduced DKL by replacing the last layer of an \ac{MLP} with a kernel layer, such as spectral mixture (SM) base kernels, and jointly learning the kernel hyperparameters along with the network weights. 
This modification combined the expressiveness of deep networks with the probabilistic properties of \acp{GP}, effectively making the network behave like a \ac{GP} with the same kernel.
Similarly, \ac{KKAN} can be interpreted as an \ac{MLP} in which the final layer is replaced by a linear combination of splines. Architecturally, both DKL and \ac{KKAN} share the idea of projecting features into the output space via a learned functional layer. However, their underlying mechanisms differ significantly. 
SM kernels in DKL architectures are composed of cosine functions operating in the frequency domain, often exhibiting periodic behavior and lacking direct spatial or semantic correspondence. 
In contrast, spline-based representations are piecewise polynomials that are well-suited to interpreting time-domain inputs. They offer higher interpretability than frequency domain representations and are particularly effective for modeling smooth, structured data. Moreover, unlike our design, DKL does not guarantee distance-awareness and may map inputs that are far apart to the same feature representation~\cite{van2021feature}. 

\noindent{\textit{}}\underline{Probabilistic distance-aware methods:}
Existing distance-aware uncertainty estimators are predominantly probabilistic, with \ac{SNGP}~\cite{liu2020simple} being a seminal example. 
In~\cite{liu2020simple}, distance-awareness is used
to quantify how far an input lies from training examples and to produce a graded measure of unfamiliarity that aids out-of-distribution (OOD) detection~\cite{yang2024generalized}. To study the distance-aware error bounds, we draw inspiration from the \ac{SNGP}, which provides formal error guarantees for \acp{NN} by leveraging Lipschitz continuity. In the \ac{SNGP} framework, weight normalization is applied to enforce a global Lipschitz constraint on the \ac{MLP}. 
Moreover, the final layer of the \ac{MLP} is replaced with a \ac{GP} layer, akin to DKL. 
However, unlike standard DKL, the \ac{GP} layer in \ac{SNGP} is implemented via random Fourier features (RFF), and the underlying approximation for tractability may lead the uncertainty to converge to zero in the data limit~\cite{van2021feature}. 
In this work, we extend these ideas in \ac{K-DAREK} by combining \acp{MLP} with a \ac{KAN}-based architecture, yielding a scalable, computationally efficient method that provides worst-case distance-aware error bounds for this network.
Unlike \ac{SNGP}, which employs a single \ac{MLP} with a \ac{GP} approximation only at the final layer, our worst-case (adversarial) approach applies a separate \ac{MLP} to each input, yielding a structurally more input-adaptive architecture. 

\Ac{DUE}~\cite{van2021feature}, similar to \ac{SNGP}, uses spectral normalization on the feature extractor and applies an approximated \ac{GP} on the final layer. Unlike SNGP, it employs an inducing-point GP instead of RFF, which yields better results; however, it may yield poor uncertainty estimates when the number of inducing points is limited or when the points do not cover parts of the feature space, especially in high dimensions.

Other probabilistic distance-aware methods include the following. \Ac{DUQ}~\cite{van2020uncertainty} is restricted to classification and models each class via a single centroid in feature space, which may be limiting. In this context, \ac{DDU}~\cite{mukhoti2023deep} fits one Gaussian mixture per class, so it assumes the features of a class follow a Gaussian distribution. 

Probabilistic distance-awareness quantifies the expected distance of a test point from the training distribution, typically using measures such as likelihood or kernel density estimates. In contrast, our worst-case (robust) distance-awareness analysis relies on deterministic bounds, defined via Lipschitz constants, to guarantee how much the output can change under any admissible perturbation.
We present a detailed empirical comparison of these probabilistic distance-aware methods with \ac{K-DAREK}, and highlight their conceptual similarities and differences. 
Also, our method is a deterministic alternative to probabilistic approaches rather than a variant of them.
In Section~\ref{sec:experiment}, we demonstrate how probabilistic models can produce misleading uncertainty estimates and overgeneralize in the absence of data.

\noindent{\textit{}}\underline{Interpretability:}
A recurring theme in this paper is interpretability.
Distance-awareness provides a form of interpretability. An estimator whose confidence reflects proximity to training data offers a human-understandable explanation for its own output, since the nearest training point and its distance explain why the output is high or low. Interpretability itself is a widely valued and inconsistently defined concept in machine learning, driven by diverse and sometimes conflicting motivations~\cite{doshi2017towards}.
Despite this ambiguity, it is common for models to be described as ``interpretable'' without precise criteria or justification~\cite{lipton2018mythos}. Different scientific disciplines offer varying intuitions about what makes a function simple or understandable~\cite{liu2025kan,marcinkevivcs2012interpretability,rudin2022interpretable}, further complicating {the pursuit of} a unified definition.
Mechanistic interpretability seeks to reverse-engineer trained models by identifying the internal components or circuits responsible for specific behaviors~\cite{nanda2023progress}, while symbolic regression focuses on discovering human-readable mathematical expressions that approximate a model’s behavior or underlying data-generating process~\cite{makke2024interpretable,tohme2024isr,kacprzyk2025beyond}.
In this paper, we define \textbf{\textit{interpretability}} as the human-understandable mapping between model inputs and outputs, particularly through closeness of predictions to training data, which intuitively supports robustness and safety~\cite{lipton2018mythos, doshi2017towards}. 
Interpretability is a subjective term which is hard to define precisely and concretely~\cite{doshi2017towards,lipton2018mythos}; here, it is connected with ``explanation by examples''~\cite{lipton2018mythos}.

\noindent{\textit{}}\underline{System verification:}
Neural network verification methods such as CROWN, $\alpha$-CROWN, $\beta$-CROWN, and auto-LiRPA~\cite{xu2020fast,liu2021algorithms,zhang2018efficient,xu2020automatic} certify output bounds for a fixed, already-trained network $\hat f$ under a specified input perturbation, by propagating bounds through its known weights.
These methods have also been applied to control-theoretic guarantees directly, for instance, certifying Lyapunov stability of neural controllers~\cite{shi2024certified}.
Our work focuses on a different problem than system verification literature.
System verification focuses on ensuring a constraint is satisfied for all inputs, $g(\hat{f}(x), x) \ge 0$ for all $x \in \calX$, given a learned function $\hat{f}(x)$~\cite{shi2024certified}, where $g$ is a task-specific constraint function (for instance, a safety or stability condition).
Typical approaches find upper and lower bounds of the learned function 
$\overline{f}(x) \ge \hat{f}(x) \ge \underline{f}(x)$, and use the bound to guarantee $\underline{g}(\hat{f}(x), x) \ge 0$.
On the other hand, we aim to find the error between a learned function $|\hat{f}(x) - f(x)|$ and an unknown function that is only observable through data samples and structural assumptions, in our case, Lipschitz continuity.
To make this comparison concrete, in Sec.~\ref{sec:experiment} (Adapted-CROWN comparison experiment), we adapt a CROWN-style linear relaxation to K-DAREK's architecture and compare the resulting output bound against K-DAREK's own error bound on the same task, anchored at the same training knots (Fig.~\ref{fig:Err-comparison}~\textbf{(d)}).
This distinction also matters directly for control applications: in our multi-agent CBF experiment (Sec.~\ref{sec:experiment}), we use K-DAREK's error bound to quantify uncertainty in a learned dynamics model, whereas system verification methods such as~\cite{shi2024certified} instead certify an upper bound on a controller's own output; the two serve different roles even when both are used inside a control pipeline.

\section{BACKGROUND}
\label{sec:background}
We begin by {defining distance-awareness} and then reviewing the key background concepts needed to understand our approach:
 a) measuring  distance-awareness~\cite{ataei2026darek}, b) \acp{KAN}~\cite{liu2025kan},  c) spectral normalization~\cite{miyato2018spectral}, and d) DAREK~\cite{ataei2025darek}.

\begin{definition}[Distance-awareness~\cite{ataei2025darek}]
\label{def:defi-dist}
An approximate function $\hat{f}(\bfx)$, defined over $\bfx \in \calX$, is \textbf{distance-aware} if its associated error or uncertainty $u_{\hat{f}}(\bfx)$ increases monotonically with the distance of a test point $\bfx$ from the training input $\calX_{\calD}$\footnote{Throughout this paper, $u_{\hat f}(x)$ 
denotes a deterministic worst-case error bound on $|f(x) - \hat f(x)|$ under the Lipschitz assumptions, as distinct from the probabilistic uncertainty (e.g., posterior variance) reported by baselines such as GP, DUE, and SNGP.}. 
\end{definition}

The distance is quantified by a function $d_u(\bfx, \bftau^*)$, which computes the minimum distance between the test point and all training samples for the uncertainty function $u$. 
We denote the nearest available training point by $\bftau^* \in \calX_\calD$. The uncertainty function $u_f$ on the direction of the distance function $d_u$ at test point $\bfx^*$ is distance-aware if
\begin{align}
    [\nabla_\bfx u_f(\bfx^*)]^\top \nabla_\bfx d_u(\bfx^*, \bftau^*) \geq 0 \quad \forall \bfx^* \in \calX.
    \label{eq:dist-aware-cond}
\end{align}

\noindent{\textit{}}\underline{Measuring distance-awareness:} 
Checking the condition in~\eqref{eq:dist-aware-cond} over the entire {continuous} input space $\calX$ is, in general, computationally intractable, as it requires establishing the global monotonicity of the uncertainty function.
The function $u_f$ is generally nonlinear. There is no closed-form way to check condition~\eqref{eq:dist-aware-cond} at every point in $\calX$.
In practice, we evaluate this property using a sampling-based approach, sampled distance-awareness (SDA)~\eqref{eq:SDA}.
This provides a practical and consistent metric for evaluating this property across different uncertainty estimators. 
By considering $d_u=\|x-\tau^*\|^2/2$ as the distance function, taking $N$ samples from the input space, and computing the frequency with which condition~\eqref{eq:dist-aware-cond} is satisfied, we arrive at a metric with the following formulation. 
Since $\nabla_x d_u(x,\tau^*) = x-\tau^*$, condition~\eqref{eq:dist-aware-cond} reduces to $[\nabla_x u_{\hat f}(x^*)]^\top(x^*-\tau^*) \ge 0$, which is exactly the sign condition evaluated inside the indicator below.
The \emph{sampled distance-awareness} (SDA) is defined as:
\begin{align}
    \label{eq:SDA}
    \text{SDA} = \frac{1}{N}\sum_{\bfx_t \sim \calX_{\text{test}}}
    \mathbbm{1}
    \left[
    \nabla_{\bfx} u_{\hat{f}}(\bfx_t)^\top (\bfx_t - \bftau^*) \ge 0 \right],
\end{align}
where $\mathbbm{1}[.]$ is the indicator function and $\bfx_t \sim \calX_{\text{test}}$ are test samples drawn uniformly from test data.
A higher SDA value indicates better distance-awareness, meaning that uncertainty is more likely to increase as the test point moves farther from the nearest training point. Our goal is therefore empirical distance-awareness, measured via SDA and reported comparatively against baselines, rather than a theoretical guarantee that Definition~\ref{def:defi-dist} holds globally. 

\noindent{\textit{}}\underline{KANs~\cite{liu2025kan}:}
\ac{KAT} states that any continuous multivariate function can be represented as a finite superposition of univariate functions and addition~\cite{schmidt2021kolmogorov}. 
Formally, a continuous multi-variate function $f: \calX \to \bbR$, ($\calX \subset \bbR^d$) can be represented as:
\begin{align} f(\bfx)=\sum_{q=1}^{2d+1}\Phi_q \big(\sum_{p=1}^{d}\psi_{p,q}(x_p)\big),
\label{eq:KAT}
\end{align}
where $\bfx \in \calX \subset \bbR^d$ is a d-dimensional input vector, and $x_p$ is its $p$-th component.  
In this formulation, $\Phi_q$ and $\psi_{p,q}$ are univariate functions corresponding to the nodes of outer and inner layers, respectively. \ac{KAN}~\cite{liu2025kan} relaxes the two-layer structure by allowing arbitrary depth and a flexible number of univariate functions. 
These functions are replaced with B-splines, leading to a hierarchical representation of a vector-value function $\bff=(f_1(\bfx),f_2(\bfx),\cdots,f_m(\bfx))$ where each component $f_r: \calX \to \bbR$ for $r \in \{1,2,\cdots,m\}$, as follows:
\begin{align}
    f_r(\bfx)&=\sum_{i_L=1}^{N_L}s_{L,r,i_{L}} \Big(\sum_{i_{L-1}=1}^{N_{L-1}}\calS_{{L-1},i_L,i_{L-1}}(\bfx)\Big), \text{where},\notag\\    
    \calS_{l,j,i}(\bfx)&=s_{l,j,i} \Big(\dots \sum_{i_2=1}^{N_2}s_{2,i_3,i_2}\big(\sum_{i_1=1}^{N_1}s_{1,i_2,i_1}(\bfx_{i_1})\big)\Big).
    \label{eq:SplineVecor}
\end{align}
Here, $s_{l,j,i}$ denotes a $k$-th order B-spline defined by its knots, located in layer $l$, projecting input $i$ to contribute to output $j$, and $N_l$ for $l \in \{1,2,\cdots, L\}$ represent the input dimension and hidden layer dimensions up to the last layer of the network. 

\noindent{\textit{}}\underline{Spectral normalization:}
Spectral normalization~\cite{liu2020simple} enforces a desired Lipschitz continuity for a single fully-connected neural network layer by normalizing its weight matrix $W$ by its spectral norm (largest singular value).
Spectral normalization is most straightforward to implement with the \ReLU{} activation function, which has a Lipschitz constant of 1.
Therefore, for a linear layer followed by a \ReLU\ activation, defined as $h_l(\bfx) = \ReLU(W_l\bfx+\bfb_l)$, the Lipschitz constant of $h_l$ can be bounded as follows:
\begin{align}
    \calL_{h_l} &= \frac{\| h_l(\bfx) - h_l(\bfx')\|}{\| \bfx - \bfx' \|}
    \le \| (W_l\bfx+\bfb_l) - (W_l\bfx'+\bfb_l) \| \notag\\
    &= \frac{\| W_l (\bfx - \bfx') \|}{\| \bfx - \bfx' \|} \le \sigma_{\text{max}}(W_l) .
\end{align}
where $\sigma_{\max}(W_l)$ is the largest singular value of the matrix $W_l$.
By normalizing the weight matrix, which constrains $\sigma_{\max}(W_l)\le \calL_{h_l}$, we ensure that each layer is Lipschitz continuous with the desired Lipschitz constant $\calL_{h_l}$. Specifically, during training, we normalize the weights of the neural network after every training iteration as described in~\cite{liu2020simple}
\begin{align}
\label{eq:spectra-MLP}
\bar{W}_l^{(t+1)} = \begin{cases}
\frac{\calL_{h_l} W_l^{(t)}}{\sigma_\text{max}(W_l^{(t)})} & \text{ if }  \calL_{h_l} < \sigma_\text{max}(W_l^{(t)})\\
W_l^{(t)} & \text{ otherwise }.
\end{cases}
\end{align}
In this paper, we refer to an \ac{MLP} with \ReLU{} activation functions in the hidden layers, a linear activation in the output layer, and scaled, spectrally normalized weights as a \ac{SNR-MLP}.

\noindent{\textit{}}\underline{DAREK~\cite{ataei2025darek}:}
Here, we review the background needed for the spline error analysis.
We denote \textit{Newton's polynomial operator} of order $k$ as defined by~\cite[p.7]{de1978practical} on a sorted set of knots $\bftau$ with $m_k$ elements by $\calP_{k,n}[\bftau,f(\bftau)](x)$. 
We use $k+1$ nearest knots to $\bftau[n]$ from $\bftau$, where $\bftau[n]$ denotes the $n$-th element of the set or vector $\bftau$, and the true output values are determined by $f(\bftau)$. 
We define the \textit{$k$-th-order Lipschitz constant}, $\calL^{k}_f$, of a $k-1$ times differentiable function $f: [a,b] \to \bbR$ as the ratio of the maximum change in the $(k-1)$-th derivative of $f$ to the change in the input,
\begin{align}
    \frac{|f^{(k-1)}(x)-f^{(k-1)}(y)|} {d(x,y)} \le \calL^{k}_f \quad \forall x \neq y.
\end{align}
A \textit{piecewise polynomial} $\hat{f}(x)$ of order $k$ on $\bftau$ is constructed by dividing the domain into $m_k-1$ intervals, each associated with a distinct polynomial segment $\hat{f}_{[j]}(x)$ of order $k$, as follows:
\begin{align}
    \label{eq:piecewise-poly}
    \hat{f}(x) &=  \sum_{i=0}^k c_{i,j}x^k \eqcolon \hat{f}_{[j]}(x) \qquad \forall x \in [\tau_j, \tau_{j+1}), \notag\\
    \text{s. t.} \quad
    \hat{f}_{[j]}(& \tau_{j+1})
    = \hat{f}_{[j+1]}(\tau_{j+1}).        
\end{align}
Here, $\hat{f}_{[j]}(x)$ denotes the $j$-th polynomial component. 
Piecewise polynomials are continuous but not necessarily smooth. 

\begin{theorem}[Interpolation error~\cite{ataei2025darek}]
\label{thm:interp-error}
For a function $f$ with $k+1$ continuous derivatives (that is, $f \in C^{(k+1)}$) and its $k+1$-th-order Lipschitz constant is $\calL^{k+1}_f$, the error at test point $x \in [\bftau[j], \bftau[j+1])$ is bounded as follows:
\begin{align}
    \label{eq:interpolation-error}
    | f(x) - \calP_{k,j}[\bftau,f(\bftau)](x) | &\le  
                \underbrace{
                \frac{\calL_f^{k+1}}{(k+1)!}
                \left|\prod_{i=j}^{j+k} (x-\bftau[i])\right|
            }_{\eqqcolon \bar{u}_f(x;\bftau)}.
\end{align}
\end{theorem}
This result assumes the approximation exactly matches the function at the knot points, $\bftau[j]$. However, in practical scenarios-such as when noise is present-exact matching at knots is generally not achievable. The following result extends the error bound to account for spline approximations that incur non-zero error at knots.

\begin{remark}
    \label{rem:tight-condition}
    The upper bound in~\eqref{eq:interpolation-error} is tight and reduces to an equality when $(k+1)$-th derivative of function $f$ is constant and equal to $\calL^{k+1}_f$, $f^{(k+1)}(x) = \calL^{k+1}_f, \quad \forall x \in [\bftau[j], \bftau[j+1]]$~\cite{ataei2026darek}.
    Hence, the bound is tighter than the trivial bound~\eqref{eq:trivial_bound_f}, for any $k+1$ Lipschitz function with constant $\calL^{k+1}$.
\end{remark}

\begin{theorem}[Spline error~\cite{ataei2025darek}]
\label{thm:spline-error}
Under the assumptions of Theorem~\ref{thm:interp-error}, consider a piecewise polynomial approximation $\hat{f}(x)$ and the error function is defined as $e_{j}^f(x) \teq f(x) - \hat{f}_{[j]}(x)$ for all $x \in [a, b]$ and has known values at the knots. Then the error for $x \in [\tau_j, \tau_{j+1})$ is bounded by,
\begin{align}
    \label{eq:spline-error}
    \hspace{-0.5em}
    |f(x)&-\hat{f}(x)| \le \bar{u}_f(x;\bftau) + |\calP_{k,j} [e_{j}^f](x)|\eqqcolon {u}_f(x;\bftau).
\end{align}
\end{theorem}

The conditions under which these error bounds are tight are detailed in~\cite{ataei2025darek}.

The error of one output of a spline layer, where $N_s$ splines contribute to the output, is computed as $u_{\spline} = \sum_{i=1}^{N_s} u_{s_i}(\bfx, \calT)$.
We also revisit the two-layer error bound presented in the DAREK paper~\cite{ataei2025darek}, as our \ac{K-DAREK} architecture adopts a two-block structure, comprising an inner \ac{MLP} block followed by a spline block.
The propagated error of a two-layer network, $\hat{f}(\bfx)=h_2(\bfh_1(\bfx))$, can be calculated using the following equation~\cite[Theorem 2]{ataei2025darek}:
\begin{align}
    \hspace{-0.38cm}
    |f(\bfx) -\hat{f}(\bfx)| \le u_{h_2}\big(\bfh_1(\bfx);\bfh_1(\calT)\big) + \calL_{h_2}^1 \mathbf{1}_{\bfh_1}^{\top} \bfu_{\bfh_1}(\bfx;\calT),
    \label{eq:two-layer-error}
\end{align}
which shows that the error from earlier layers is scaled by the Lipschitz constant of the current layer, $\calL_{h_2}^1$, and then added to the current layer's error. Here, $\mathbf{1}_{\bfh_1}$ is a vector of all ones with the same size as $\bfh_1$. In~\eqref{eq:two-layer-error} $\calT$ is a matrix of knots of size $N_s \times m_k$, where $m_k$ defines knots for each of the $N_s$ splines in layer $h_1$. This two-layer error propagation has been extended to arbitrarily deep spline networks~\cite{ataei2025darek}.

\begin{remark}
    Theorem~\ref{thm:interp-error} follows from the classical Newton interpolation remainder formula. When a degree-k polynomial exactly interpolates the function at the knot points, the approximation error between knots is entirely determined by the function's $(k+1)$-th-order variation. Bounding this variation using the $(k+1)$-th-order Lipschitz constant yields the computable worst-case error bound in~\eqref{eq:interpolation-error}, which is zero at the knots and grows smoothly between them. Theorem~\ref{thm:spline-error} extends this result to the practical setting where the approximation does not exactly match the function at the knots. In this case, the total error naturally decomposes into two components: the interpolation error characterized by Theorem~\ref{thm:interp-error} and the error introduced by the residuals at the knot values. By the linearity of polynomial interpolation and the triangle inequality, these two contributions combine to produce the bound in~\eqref{eq:spline-error}. 
    Equation~\eqref{eq:two-layer-error} is obtained similarly: writing $|f(x) - \hat f(x)| = |h_2(h_1(x)) - \hat h_2(\hat h_1(x))|$, we insert and subtract $h_2(\hat h_1(x))$, and bound the two resulting terms via the triangle inequality, so that propagation is governed by the Lipschitz constant of the outer function $h_2$.
\end{remark}

Next, we introduce the hybrid \ac{K-DAREK} architecture, which integrates \ac{MLP} and spline components, and develop its error bound using Lipschitz continuity.

\section{PROBLEM FORMULATION AND OVERVIEW OF THE RESULTS}
\label{sec:problem-and-overview}
\begin{figure*}[ht]
  \centering
  \includegraphics[width=0.60\textwidth,trim=0 0 0 0]{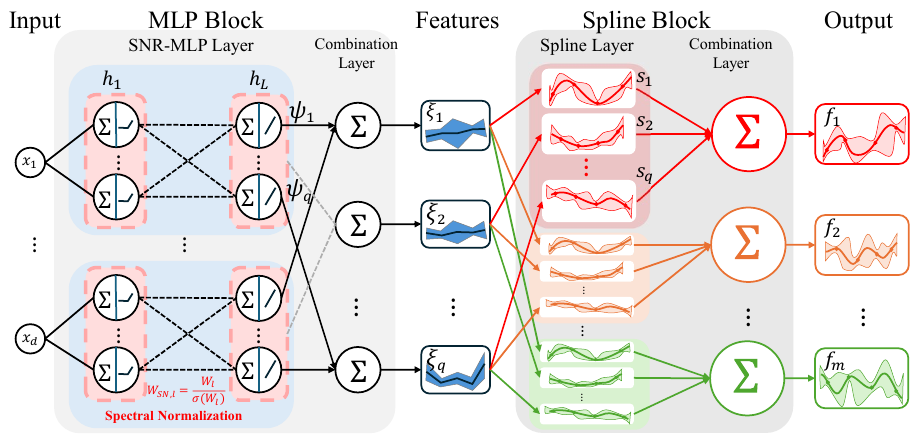}%
  \hspace{1.0cm}
  \includegraphics[width=0.234\textwidth,trim=0 0 0 0]{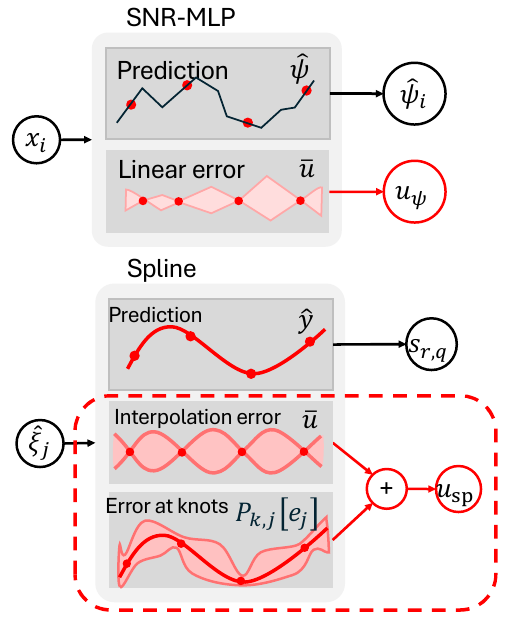}
    
  \caption{
  \textbf{left)} \Ac{K-DAREK} architecture consists of two blocks: \ac{MLP} and spline.
  The \ac{MLP} block contains one \acs{SNR-MLP} for each input dimension. A combination layer unifies the output of \acsp{SNR-MLP} into a feature vector. The spline block consists of a Spline Layer followed by a combination layer that constructs the model's final output.
  \textbf{right)} The error analysis of \ac{K-DAREK} has two main components. {A single \ac{SNR-MLP} receives input $\bfx_i$ and computes a feature, $\hat \psi_i$, along with the linear error~\eqref{eq:MLP-error}. A single Spline employs features $\hat \xi_j$ to calculate the spline values $s_{r,q}$ and spline error using interpolation error and error at the knot as described in~\eqref{eq:two-layer-error}.}}  
  \vspace{-3mm}
  \label{fig:K-DAREK}
\end{figure*}

The applicability of \Acp{NN} is limited in the absence of a measure of uncertainty or error, particularly in safety-critical settings. 
In this section, we formulate the problem of deriving distance-aware worst-case error bounds for \acp{NN} and discuss their role in safe control and other applications.

A desirable property of an uncertainty estimate is distance-awareness: it should increase monotonically as the input moves away from the training data. We formalize worst-case distance-aware uncertainty as follows.

To characterize worst-case uncertainty, let $f: \calX \to \bbR$ ($\calX \subset \bbR^d$) be an unknown target function, and let $\calF$ be the admissible function class, which encodes the structural properties that we expect the target
function $f \in \calF$ to satisfy. 
A dataset $\calD = \{(\bfx_i, y_i)\}_{i=1}^{n}$ is observed, {where} $y_i = f(\bfx_i) + \eta_i$,  
where $\eta_i$ is observation noise that satisfies $|\eta_i| \leq \epsilon_{\max}$, where $\epsilon_{\max} \geq 0$\footnote{Since our analysis is worst-case, we assume bounded noise. However, this framework naturally extends to the probabilistic setting: if the noise can be bounded with high probability, the deterministic ambiguity set~\eqref{def:ambigu} generalizes to a probabilistic feasible set~\cite{bertsimas2021probabilistic,wang2023learning}, $\{ g \in \calF \mid P( \ell(g(\bfx_i), y_i) \le \epsilon_{\text{th}}) \geq 1-\beta, \, \forall (\bfx_i, y_i) \in \calD \}$, where the constraint is required to hold with probability at least $1-\beta$ rather than deterministically.}.

Given this dataset and a tolerance parameter $\epsilon \geq 0$, we define the following ambiguity set
\begin{align}
\label{def:ambigu}
    \calF_\calD \coloneq \{g \in \calF | \, \ell(g(\bfx_i), y_i) \leq \epsilon, \forall (\bfx_i, y_i) \in \calD \},
\end{align}
where $\ell$ is a discrepancy measure. $\calF_\calD$ is a subset of admissible functions consistent with the data given the tolerance parameter $\epsilon$. By choosing a larger $\epsilon$, the ambiguity set $\calF_\calD$ enlarges, leading to more conservative bounds, as described next.

Consider a predictor $\hat{f}_{\Theta^*}(\bfx)$ that can, for instance, be trained on $\calD$ by minimizing
\begin{align}
    \Theta^* = \argmin_{\Theta} \frac{1}{|\calD|}\sum_{(\bfx_i,y_i) \in \calD} \ell(\hat{f}_{\Theta}(\bfx_i),y_i),
\end{align}
where $\Theta^*$ are optimized parameters of the model, and $|\calD|$ denotes the size of the data set. Then the associated worst-case uncertainty bound (prediction error) at a test point $\bfx^*$ is
\begin{align}
    J_{\epsilon}(\bfx^*) \coloneq \sup_{g \in \calF_\calD} \ell(g(\bfx^*), \hat{f}_{\Theta^*}(\bfx^*)).
\label{eq:worst_case_uncertainty}
\end{align}
This quantity captures the worst-case deviation across all admissible target functions compatible with the observed data, up to the tolerance parameter $\epsilon$.
Determining $J_{\epsilon}(\bfx^*)$ is generally intractable, as it involves optimization over an infinite-dimensional function space, and in this work, we suitably restrict the class $\calF$, as discussed next.

To formalize the problem, we use the definition of distance-awareness given in Sec.~\ref{sec:background} (Definition~\ref{def:defi-dist}). 

Having established the worst-case error bound~\eqref{eq:worst_case_uncertainty} and formalized the notion of distance-awareness, we now state the main problem. 

\begin{problem}
\label{problem-state}
Given a dataset $\calD$ and a trained model $\hat{f}$, construct an upper bound $u_f(\bfx; \calD)$ on the prediction error,
\[|g(\bfx)-\hat{f}(\bfx)| \leq u_f(\bfx; \calD), \quad \forall \bfx \in \calX \And \forall g \in \calF.\]
The bound needs to satisfy four properties:

(i) Validity---the bound holds for every function $g$ in the admissible class $\calF$  defined in~\eqref{def:ambigu};  (ii) Distance-awareness---$u_f$ satisfies Definition~\ref{def:defi-dist}; (iii) Tightness---$u_f$ should  neither substantially overestimate the true error nor underestimate it---{defined as the functional $\rho(u_f) \defeq\int_{\calX} [u_f(x;\calD) - J_{\epsilon}(x)] dx$, which we seek to keep small}; (iv)  Efficiency---$u_f$ can be efficiently evaluated {in terms of computational costs.}
\end{problem}
Solving this problem is intractable, as it requires optimization over an infinite-dimensional function space. 
In this work, to take steps toward addressing the above problem, we restrict $\calF$ to the class of Lipschitz continuous functions with known constant $\calL_f$. 
If we assume both true and learned functions $\hat{f}$ have the same Lipschitz $\calL_f$, then a trivial upper bound for~\eqref{eq:worst_case_uncertainty} will be:
\begin{align}
u^{triv}_f(\bfx^*;\calD) = 2 \calL_f \|\bfx^* - \bftau^*\| +  \epsilon_{\max} \ge J_\epsilon(\bfx^*
),
\label{eq:trivial_bound_f}
\end{align}
where $\bftau^*$ is the nearest training point to the test point, 
\begin{align}
    \bftau^* = \argmin_{\bfx \in \calX_\calD } \|\bfx^*-\bfx\|, 
    \label{eq:nearest_training_point}
\end{align}
where $\calX_\calD = \{\bfx_i:(\bfx_i,\bfy_i) \in \calD\}$.
 
Using the distance function $d_u = \|\bfx^* - \bftau^*\|$, the trivial bound~\eqref{eq:trivial_bound_f} is distance-aware because it depends on $\bfx^*$  and it is monotonically increasing in this distance.
Particularly, the left-hand side of~\eqref{eq:dist-aware-cond} is equal to $2 \calL_f \| \nabla_\bfx d_u \|^2$, which is non-negative and holds for any $\bfx^*$.

To obtain an efficient and tight bound in \acp{NN}, we leverage the function's Lipschitz continuity and propagate the error layer by layer through the model's structure. 
For simplicity, consider a two-stage composition $\hat{f}(\bfx) = \hat{h}_2 \circ \hat{h}_1(\bfx)$, where the Lipschitz constants of $\hat{h}_1$ and $\hat{h}_2$ are  $\calL_1$ and $\calL_2$, respectively, and $\calL_f \leq \calL_1 \calL_2$. 
Also, let $u_{h_1}(\bfx) \geq \|h_1(\bfx) - \hat{h}_1(\bfx)\|$ and $u_{h_2}(\bfz) \geq \|h_2(\bfz) - \hat{h}_2(\bfz)\|$ be the per-block error bounds.
Then the total error can be decomposed into layer-wise errors:
\begin{align}
    |f(\bfx)-\hat{f}(\bfx)| \leq u_{h_2}(\hat{h}_1(\bfx)) + \calL_2 u_{h_1}(\bfx).
    \label{eq:two_block_bound}
\end{align}
This proposed error bound~\eqref{eq:two_block_bound} is tighter than the trivial bound in~\eqref{eq:trivial_bound_f} when the per-block error estimates $u_{h_1}$, $u_{h_2}$ are tighter than their respective trivial bounds
(see Remark~\ref{rem:tight-condition} for more details).
Note that in Equation~\eqref{eq:two_block_bound}, without loss of generality, $\epsilon_{\max}$ is implicitly considered inside $u_{h}$. In the proposed framework, its effect is recovered through the error-at-knots term in the spline error analysis.

Specifically, in this paper, we adopt a \ac{KKAN}-style architecture, 
\[\hat{f}(\bfx) = \hat{h}_\spline \circ \hat{h}_\MLP(\bfx), \]
where the \ac{MLP} has Lipschitz constant $\calL_\MLP$, and the spline layer has Lipschitz constant $\calL_\spline$. The model architecture consists of two building blocks: \ac{MLP} block and spline block, as shown in Fig.~\ref{fig:K-DAREK}. In the MLP block, each input dimension is transformed by a spectrally normalized MLP layer that enforces a known Lipschitz constant. The resulting representations are summed to form a unified feature vector. The spline block then maps this feature vector to the output via splines, whose interpolation error can be bounded analytically.

As discussed next, while this approach does not fully solve Problem~\ref{problem-state}, {our goal is a tractable worst-case bound that} advances beyond every state-of-the-art baseline on at least one key criterion. 

\subsection{Overview of Applications}
\label{sec:overview}
\begin{figure}[ht]
  \centering
  \includegraphics[width=0.45\textwidth, trim=0 0 0 0, clip]{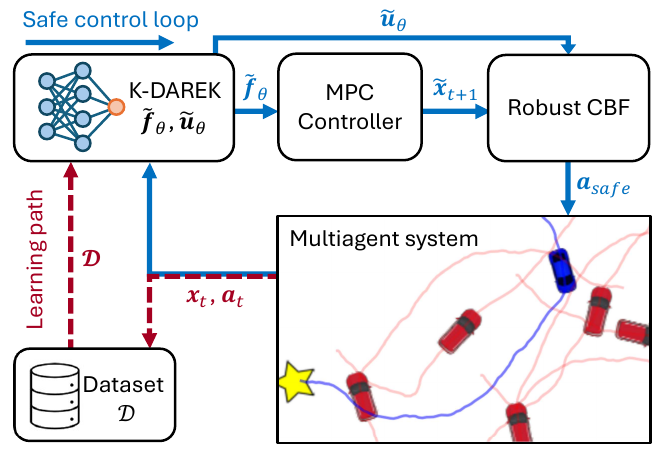}
  \caption{
  Multi-agent safe control diagram. \red{Red arrows $\mathbf{\to}$} show the learning path. 
  In this path, states $\bfx_t$ and controls $\bfa_t$ are collected as inputs and $\bfx_{t+1}$ as outputs in the dataset $\calD$, and K-DAREK is trained to predict the system dynamics~\eqref{eq:system-dynamics}. 
  \blue{Blue arrows $\mathbf{\to}$} show the safe control loop. 
  It passes learned dynamics $\Tilde{\bff}_{\theta}$ and estimated uncertainty $\Tilde{\bfu}$ along with the nominal trajectory from the MPC controller $\Tilde{x}_{t+1}$ to a robust CBF to produce the safe control $\bfa_{\text{safe}}$ and applies it to the controlled agent. 
  The multi-agent plot in the diagram shows one of the successful trajectories using \ac{K-DAREK} bounds, where the controlled agent (\blue{blue car}) safely navigates to the goal location while avoiding the other agents (\red{red cars}), see Fig.~\ref{fig:multiagent-traj}.
  }
  \label{fig:Safe-Control}
\end{figure}

We validate our approach in both control-theoretic and machine learning applications, as described next.
\noindent{\textit{}}\underline{Safe navigation in unstructured environments:} 
To validate \ac{K-DAREK} in a safety-critical setting, we evaluate it on a multi-agent navigation task in an unstructured environment. In this scenario, a controlled agent must navigate from a start position to a goal while avoiding collisions with other agents that follow unknown and heterogeneous control policies~\cite{cheng2020safe}. 
The control architecture is two-layered: an \ac{MPC} controller first computes a goal-directed but potentially unsafe trajectory, and a \ac{CBF}-based safety filter then solves a quadratic program to find the closest safe control input that respects collision avoidance constraints, as depicted in Fig.~\ref{fig:Safe-Control}. The key challenge is that the \ac{CBF} requires bounds on the uncertain dynamics of agents. 
In~\cite{cheng2020safe}, these bounds were obtained using a matrix-variate \ac{GP}~\cite{louizos2016structured}, which becomes computationally prohibitive as data grows. In this work, \ac{K-DAREK} replaces the \ac{GP} with distance-aware error bounds from the \ac{KKAN} architecture, demonstrating that the framework {preserves the structure of the original CBF safety condition, conditional on the validity of the assumed error bound and Lipschitz constant}
% retains the safety guarantees of the original \ac{CBF} formulation 
at a fraction of the computational cost.

\noindent{\textit{}}\underline{Other machine learning applications:} Beyond the safe control application, we evaluate \ac{K-DAREK} on  real-world regression tasks (e.g., Real Estate Valuation~\cite{frank2010uci}) and Celebrity Attribute face detection
dataset~\cite{liu2015deep} and assess its uncertainty quantification against  state-of-the-art approaches, including GP-based~\cite{rasmussen2006gaussian, hensman2015scalable}, deep uncertainty methods~\cite{lakshminarayanan2017simple,liu2020simple, van2021feature}, and worst-case analysis methods~\cite{ataei2025darek}. Across these benchmarks, we report two key metrics: the violation rate, which measures how often the true function value falls outside the predicted error bound, and sampled distance-awareness, which quantifies whether the model's uncertainty grows appropriately as test points move farther from training data.
K-DAREK exhibits strong {empirical} distance-aware behavior, tight {error bounds,}
% uncertainty quantification, 
low \ac{MSE}, and computational efficiency compared to baselines.

\noindent{\textit{}}\underline{Key takeaways:}
We compare the proposed K-DAREK method against state-of-the-art uncertainty methods, including \ac{SNGP}, \ac{DUE}, \ac{GP}, and ensemble of \acp{KAN}, across multiple metrics: training and inference process time, memory usage, error bound violation rate, \ac{MSE}, and sampled distance-awareness. 
While K-DAREK does not dominate all baselines across every metric simultaneously, our results show that it consistently \textbf{outperforms each baseline in at least one of these metrics} (and often in several) [cf. Tables~\ref{tbl:fixed_params_budget},~\ref{tbl:real_data_allfeatures}, and \ref{tbl:facerecognition}]. 
In particular, when \ac{NN} approaches such as \ac{DUE} and \ac{SNGP} achieve competitive prediction accuracy, \ac{K-DAREK} offers {empirically} stronger distance-aware behavior {(higher SDA)} and a lower error bound violation rate; conversely, when \ac{GP}-based methods provide principled uncertainty bounds, \ac{K-DAREK} scales significantly more favorably with dataset size. Taken together, these results demonstrate meaningful progress toward solving Problem~\ref{problem-state}, advancing beyond existing methods even if a complete solution remains open. 
{In this sense, K-DAREK represents a tractable, efficiently computable estimate of the worst-case bound.
}

\section{ARCHITECTURE AND METHOD}
\label{sec:method}
This section details the \ac{K-DAREK} architecture and presents our approach for computing its worst-case error bound.

Building upon \ac{KAN}, \ac{KKAN}~\cite{toscano2025kkans} insists on the two-layer structure of \ac{KAT}~\eqref{eq:KAT} as a composition of two functional blocks, where $\psi$ represents the inner block and $\Phi$ the outer block. The \ac{MLP}s in the inner block expand each input dimension into a higher-dimensional space through basis functions, such as Chebyshev basis functions, project this representation into a hidden space using the \ac{MLP}, and finally map it to the output space via a combined output layer. 
Intuitively, the KKAN inner block processes each scalar input coordinate independently in three steps: (1) lift the scalar $x_p$ to a short vector of Chebyshev polynomial features $\bfT_n(x_p)$; (2) pass this feature vector through a small MLP to produce a hidden-space representation; (3) sum these per-dimension representations and map the result through the spline block to the output.
For an input $\bfx \in \bbR^d$, intermediate feature dimension $q$, and an output dimension $r$, the structure of \ac{KKAN} can be written as: 
\begin{align}
    \hspace{-0.4cm}
    f_r(\bfx) = \sum_{i_s=1}^{q}s_{r,i_s}
    \Big(\sum_{p=1}^{d}
    \underbrace{\mathbf{1}_{on}^\top \bfT_{n}\big(\MLP_{p,i_s,o}(\bfT_n(x_p))\big)}_{\psi_{p,i_s}(x_p)}\Big),
    \label{eq:kkan}
\end{align}
where $\bfT_n(\bfy) \defeq (T_{0}(\bfy), \dots, T_{n-1}(\bfy)) \in \bbR^{on}$ is the Chebyshev basis operator that expands each dimension of an input vector $\bfy \in \bbR^o$ to the first $n$ Chebyshev polynomials of the first kind, $T_{n}(\bfy) \defeq \cos(n \arccos(\bfy)) \in \bbR^o$. We assume that $T_n(\bfy)$ applies element-wise when operating on a vector. Also, $s_{r,i_s}$ represents the spline that maps the $i_s$-th dimension of the intermediate layer to the output dimension $f_r$. $\MLP_{p,i_s,o} : \bbR^{n} \to \bbR^o$ is an \ac{MLP} that produces a vector output of a desired dimension $o$. 

In \text{K-DAREK}, we introduce two modifications to \ac{KKAN}. First, we restrict the Chebyshev operator to the first two degrees of Chebyshev functions, $T_0(\bfy) = 1$ and $T_1(\bfy) = \bfy$.
This simplification makes error analysis more efficient at the cost of the model's expressiveness, as it removes the nonlinear contribution of the higher-order Chebyshev terms. 
Second, we use a spectrally normalized \ac{MLP} (\ac{SNR-MLP}) instead of $\MLP_{p,i_s,o}$ in \eqref{eq:kkan}, which allows us to control the Lipschitz constant of the \ac{MLP}.

The proposed architecture, as illustrated in Fig.~\ref{fig:K-DAREK}, consists of two components: \textbf{left)} the main model architecture and \textbf{right)} the error analysis framework.
\subsection{MODEL ARCHITECTURE}
The model architecture consists of two building blocks: \ac{MLP} block and spline block, as shown in Fig.~\ref{fig:K-DAREK}. 

\noindent{\textit{}}\underline{MLP Block:}
There are $d$ \acp{SNR-MLP} followed by a combination layer.  
Each \ac{SNR-MLP} processes a single input component $x_p$ independently and maps it to a feature map $\bfpsi_{p}(x_p) = [\psi_{p,i_s}(x_p)]_{i_s=1}^q\in \bbR^q$ (see \eqref{eq:kkan}). 
The combination layer then aggregates $\bfpsi_p(x_p)$ across the input dimensions to produce the final feature vector $\bfxi = \sum_{p=1}^d \bfpsi_p(x_p) \in \bbR^q$. 

\noindent{\textit{}}\underline{Spline Block:}
As shown in Fig.~\ref{fig:K-DAREK}~\textbf{(left)} and \eqref{eq:kkan}, the Spline Layer in this block consists of $m$ groups, each containing $q$ univariate splines that produce outputs $\bfs_r=(s_{r,1},s_{r,2},\cdots,s_{r,q})$. 
The combination layer then aggregates each group by summing its spline outputs, resulting in a vector $\bff = ( f_1, \dots, f_m )= (\sum_{i_s=1}^q s_{r,i_s})_{r=1}^m \in \bbR^m$.

\subsection{ERROR ANALYSIS FRAMEWORK FOR K-DAREK}
\label{sec:error-analysis}
We divide the error analysis framework into five components described in this subsection.
At a high level, computing the error bound at a test point $x^*$ proceeds in three stages, mirrored in Alg.~\ref{alg:K-DAREK}: (1) pass $x^*$ through the MLP block to get its feature vector and the MLP block's error via~\eqref{eq:MLP-error}; (2) for each spline feature, locate its knot interval by binary search, fit a Newton polynomial to the nearest knots using the errors already observed there during training, and evaluate the local spline error via~\eqref{eq:nearest_training_point}; (3) propagate the MLP error through the spline block's Lipschitz constant and add it to the spline error via~\eqref{eq:K-DAREK}. 
%----------------- Algorithm --------------%
\begin{algorithm}
\small
\algorithmfootnote{$E$ is matrix of size $Y$ and $\bfe_i$ is $i$-th column of $E$. \\
$\bfkappa_n[j]$ is $j$-th element of $n$-th column of $K$.}
\DontPrintSemicolon
\KwData{Input-feature-output tuple $\{\calT, K, Y\}$, trained model $\hat{f}$, test point $\bfx^*$, order $k$, Lipschitz constants $\calL^{(k+1)}_f$, $\calL^{(1)}_f$.}
Precompute errors $E^f = |Y - \hat{f}(\calT)|$ \;
Divide $\calL_f^{(1)}$ equally among blocks ($\calL_{\MLP}=\calL_{\spline}=\sqrt{\calL_f^{(1)}}$).\;
Find feature of test input $\bfpsi^* = \Psi(\bfx^*)$ \; 
Find $\calL_{\overline{\spline}} = \calL_{\spline} / q$, $\calL_{\overline{\spline}}^{(k+1)} = \calL_{f}^{(k+1)} / q$, and $\calL_{\overline{\MLP}} = \calL_{\MLP} / d$ \; 
% \tcc{$\calO(\log_2(m_{k}) d)$ binary search.}
Find $u_{\MLP}$ using [Eq.~\eqref{eq:MLP-error}] \qquad {\tcc{$\calO(\log_2(m_{k}) d)$}}
\For{$n \in \{1, \dots, q\}$}{
    \tcc{$\calO(\log_2(m_{k}))$ binary search.}
    Find $j$ such that $\bfkappa_{n}[j] \le \bfpsi_n < \bfkappa_{n}[j+1].$  \;
    Fit Newton's Polynomials $\calP_{k,j}[\bfe_n^{f}]$ on the knots $\{(\bfkappa_{n}[i], E^{f}_{i})\}_{i=j}^{j+k}$\;
    Find $u_{\spline_n}(x_n^*; \bftau_n)$ [Eq.~\eqref{eq:spline-error}].\;
    }
% \tcc{$\calO(L m_h^2)$ forward pass.}
Compute error bound over $f$ [Eq.~\eqref{eq:K-DAREK}] {\tcc{$\calO(L m_h^2) + \calO(q m_k)$}}
\caption{K-DAREK
\label{alg:K-DAREK}}
\end{algorithm}

\begin{figure*}    
    \includegraphics    [width=0.98\textwidth,trim=0 0 0 0]{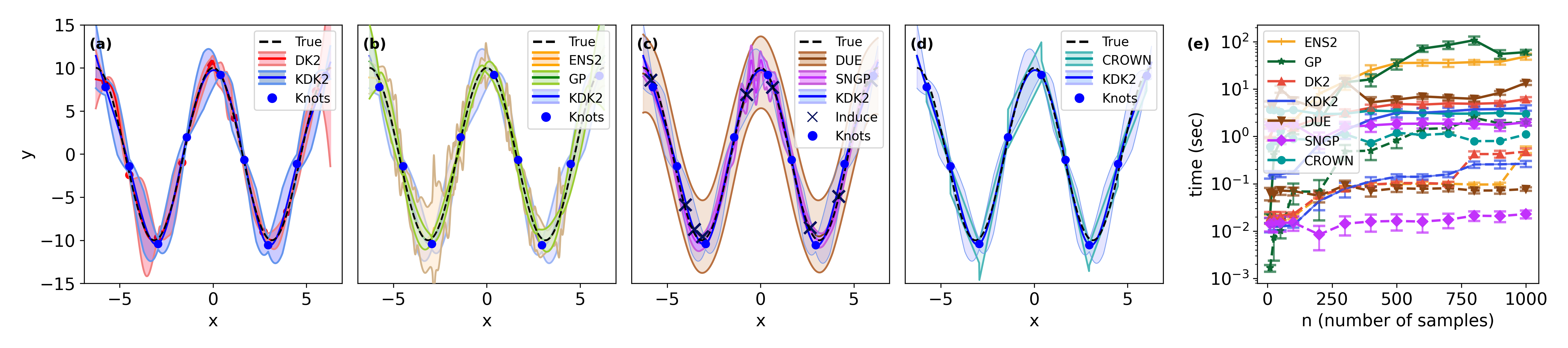}    
    \caption{Cosine function approximation experiment {(qualitative comparison)}. \textbf{(a)} Error estimation comparison of DK2 and KDK2, showing that KDK2 achieves smoother function fitting without requiring an explicit regularizer. In both models, the error bounds increase as test points move farther from the nearest knot, reflecting distance-aware error behavior. 
    \textbf{(b)} $3\sigma$ uncertainty bounds comparison between ENS2 and a \ac{GP}. The ENS2 uncertainty bounds lack distance-awareness and do not follow a consistent spatial pattern, in contrast to DK2, KDK2, and \ac{GP}, {each shown alongside KDK2 for direct comparison}.
    \textbf{(c)} $3\sigma$ uncertainty bounds comparison between \acs{DUE} and \acs{SNGP}{, shown alongside KDK2 for direct comparison}. Inducing points are projected back into the input domain and plotted as $\times$. Both uncertainties are almost uniformly distributed across the input domain in this sparse dataset.
    {\textbf{(d)}
    Comparison of the KDK2 error bound (blue) with the certified output bounds computed using forward CROWN through \texttt{auto\_LiRPA}. 
    The learned model for CROWN is same KDK2 and intervals are defined between knots. CROWN certifies the model output over each specified perturbation region, whereas KDK2 estimates an error bound based on the distance from the nearest training knot.}
    \textbf{(e)} Process time comparison of \blue{KDK2 (blue)}, \red{DK2 (red)}, {\color{darkgreen}\ac{GP} (darkgreen)}, \orange{ENS2 (orange)}, {\color{darkbrown}\acs{DUE} (brown)}, {\color{lightpurple}\acs{SNGP} (purple)}, and {\color{cyan} CROWN (cyan)} models. Training time (solid line) and inference time (dashed). The error bars indicate the variability observed across repeated runs of the experiment. 
    }
    \label{fig:Err-comparison}
\end{figure*}

\noindent{\textit{}}\underline{1) Knot Selection:}
Each spline in Spline Layer requires $m_k$ knots, and each group contains $q$ univariate splines, forming a knot matrix $K \in \bbR^{m_k \times q}$, where column $c$ corresponds to the knots of the $c$-th spline in each group, $\bfkappa_c$. 
To construct this matrix, we select $m_k$ samples from the training data during network training, forming a data matrix $\calT \in \bbR^{m_k \times d}$, where column $c$ represents the $c$-th dimension of the selected samples, denoted by $\bftau_c$.
Corresponding to these input samples, we define a label matrix $Y \in \bbR^{m_k \times m}$, where column $c$ contains the true target values for the $c$-th output dimension, $\bfy_c$. 
The knots serve as representatives of the training data, and we use the actual error at the knot locations to compute the \textit{error at knots}, {which contributes to the overall error bound}. Note that matrices $\calT$ and $Y$ are paired samples selected from the training data, and to compute the knot matrix $K$, we pass $\calT$ through the network and collect the resulting features. 

\begin{remark}
    \label{rem:knot-sel}
    Choosing $m_k$ samples from the training data influences the training and the downstream error bound, since the knots determine both the support of the spline and the reference points for the error-at-knots term.
    Several selection strategies are possible, including uniform random sampling, k-means, weighted k-means~\cite{de2012minkowski}, Chebyshev points~\cite{wahba1990spline}, and Latin hypercube sampling~\cite{mckay2000comparison}, all of which are implemented in the accompanying codebase. 
    For the results reported in this paper, we use k-means because it places knots in regions where training data is present. 
    % A systematic comparison of these strategies and their effect on bound tightness and distance-awareness is left for future work. 
    {We compare these strategies in the ablation study (Table~\ref{tbl:ablation}); a more exhaustive comparison of their effect on bound tightness and distance-awareness across additional tasks and datasets is left for future work.}
\end{remark}

\noindent{\textit{}}\underline{2) Error of MLP Block:}
MLP block has $d$ \acp{SNR-MLP}, one per input dimension, and each \ac{SNR-MLP} has $L$ layers. 
We denote the Lipschitz constant of the block by $\calL_{\MLP}$ and each \ac{SNR-MLP} by $\calL_{\MLP_i}$, with $\calL_{\MLP} \leq \sum_{i=1}^L \calL_{\MLP_i}$.
Given a test point $\bfx^*$, the error of the $i$-th \ac{SNR-MLP} is
\begin{align}
    u_{\MLP_i}(x^*_i,\bftau_i) &= \calL_{\MLP_i} \hspace{0.6mm} \|x^*_i - \tau^*_i\|,
    \notag\\
    \text{where} \quad
    \tau^*_i &= \argmin_{\tau^\dagger \in \bftau_i} \| \tau^\dagger - x^*_i  \|.
    \label{eq:SNR-MLP-error}
\end{align}
Here, $x^*_i$ is the $i$-th component of the test input  $\bfx^*$, and $\tau^*_i$ is the $i$-th component of the closest sampled data to $x^*_i$.
We obtain the worst error bound for the \ac{MLP} block by summing the individual bounds across all input dimensions:
\begin{align}
    \hspace{-0.3cm}
    u_{\MLP}(\bfx^*,\calT) &= \sum_{i=1}^{d} u_{\MLP_i}(x^*_i,\bftau_i) = \sum_{i=1}^{d}\calL_{\MLP_i} \hspace{0.6mm} \|x^*_i - \tau^*_i\|. 
\end{align}%
If all \acp{SNR-MLP} have equal Lipschitz constant $\calL_{\overline{\MLP}}$ (so that $\calL_{\MLP} \leq d\, \calL_{\overline{\MLP}}$), this expression simplifies to
\begin{align}
    u_{\MLP}(\bfx^*,\calT) &= \calL_{\overline{\MLP}} \sum_{i=1}^{d} \hspace{0.6mm} \|x^*_i - \tau^*_i\|. 
    \label{eq:MLP-error}
\end{align}
Note that each $\tau^*_i$ may originate from a different training sample for different input dimensions, so further simplification beyond this equation is not possible.

\noindent{\textit{}}\underline{3) Error of Spline Block:}
Denote a spline in the Spline Layer by $\spline_{r,i}$, where $r$ is the output index and $i$ is the feature index. Then the set of knots of this spline is $\bftau_i$, and the corresponding output values required for error calculation are $\bfy_r$.
Given this, for a test point $\bfx^*$ and its corresponding feature $\bfxi^*$, the spline error in \ac{K-DAREK} framework is defined as:
\begin{align} 
    &\hspace{-0.3cm}
    u_{\spline_{r,i}}(\xi^*_i,\bfkappa_i) \coloneqq \bar{u}_{\spline_{r,i}}(\xi^*_i;\bfkappa_i)     
    + \big\| \calP[(\frac{\hat{f}_r(\bftau)-\bfy_r}{q})_{[j]}^{\spline}](\xi^*_i) \big\|.
    \label{eq:K-DAREK-spline-error-one-sp}
\end{align}
The factor $1/q$ comes from the equal sharing of error across the $q$ splines within each group, as discussed below under Lipschitz and error sharing.
The worst error bound of the $r$-th group of splines is given by:
\begin{align}    
    u_{\spline_{r}}(\bfxi^*,K) &\coloneqq \sum_{i=1}^{q} u_{\spline_{r,i}}(\xi^*_i;\bfkappa_i).
    \label{eq:K-DAREK-spline-error}
\end{align}

\noindent{\textit{}}\underline{4) Total Error:}
Having established the error for both the \ac{MLP} and spline blocks in Equations~\eqref{eq:MLP-error} and \eqref{eq:K-DAREK-spline-error}, we can now compute the overall error of \ac{K-DAREK}. The total error of $r$-th component of $f$ at a test point $\bfx^*$ is given by:
\begin{align}
    u_{f_r}(\bfx^*) = u_{\spline_r}(\bfxi^*, K) + \calL_{\spline}^{1} u_{\MLP}(\bfx^*,\calT).
    \label{eq:K-DAREK}
\end{align}
Here, $\calL_\spline^1$ denotes the Lipschitz constant of the spline block, which propagates the uncertainty from the \ac{MLP} block into the spline block. This formulation accounts for both interpolation and transformation errors within the model architecture.

\noindent{\textit{}}\underline{5) Lipschitz and Error Sharing:}
To use the provided error analysis and ensure that the neural network preserves the stability and bounded sensitivity of the target function $f$, the approximated model must exhibit Lipschitz continuity consistent with the Lipschitz constant of $\calL_f$.
To achieve this, inspired by~\cite{liu2020simple}, we divide the Lipschitz constant of the true function equally between the \ac{MLP} and spline block $\calL_{\MLP} = \calL_{\spline} = \sqrt{\calL_f}$.
We divide the Lipschitz constant between \ac{SNR-MLP} components equally $\calL_{\overline{\MLP}} = \calL_{\MLP} / d$, and also, we apply the Lipschitz constant of $\calL_l = \sqrt[L]{\calL_{\overline{\MLP}}}$ using spectral normalization on each layer of these components.
Also, we assign an equal Lipschitz constant for each spline in the spline block as $\calL_{\overline{\spline}} = \calL_{\spline} / q$.

Additionally, we assume the error at knots arises solely from the spline block~\cite{ataei2026darek}, so we divide it equally between splines $(\hat{f}_r(\bftau)-\bfy_r)/q$.

\begin{remark}
\label{rem:est-lip}
The worst-case guarantee established above~\eqref{eq:K-DAREK} holds under the assumption that $\calL_f$ is a valid upper bound on the true Lipschitz constant of $f$. 
In practice, $\calL_f$ is estimated numerically from the training data rather than known exactly. 
\end{remark}

\begin{remark}
    \label{rem:lip-div}
    Allocating Lipschitz constants and error divisions are challenging problems, and several approaches, including heuristic allocation and optimization-based methods, can be used for this purpose. However, these questions are not the focus of this paper and have been explored in the \ac{DAREK} paper~\cite{ataei2026darek}. We defer a more detailed study of these options for \ac{K-DAREK} to future work.
\end{remark}

The key theoretical difference between DAREK (built on KAN) and K-DAREK (built on KKAN) lies in the depth and compositional structure of the spline sub-network.
DAREK stacks spline layers homogeneously to arbitrary depth, Eq.~\eqref{eq:SplineVecor}, whereas K-DAREK replaces most of that stack with a single dense (MLP) block. 
In DAREK, the worst-case error must be propagated recursively through $L$ nested spline compositions, with each layer's Lipschitz constant scaling all upstream error terms. 
K-DAREK instead has only two blocks, so its error propagation requires applying the two-layer bound~\eqref{eq:K-DAREK} only once.
Additionally, it enforces a Lipschitz constant of the MLP block through spectral normalization.
Consequently, K-DAREK's bound is less sensitive to Lipschitz division and error sharing.

\subsection{ALGORITHM COMPLEXITY}
\label{sec:algorthimic_complexity}
The computational complexity of the proposed and baseline models varies based on their architectures.
For an ensemble of $n_e$  \acp{NN} with $L$ layers and $m_h$ hidden units per layer, the complexity of inference scales as $\calO(n_e L m_h^2)$.
In contrast, a \ac{GP} depends on the number of training samples $n$ and has a training computational complexity of $\calO(n^3)$, and the inference complexity is  $\calO(n^2)$. 
DAREK model is composed of $L$ layers with $m_h$ hidden units per layer and $m_k$ knots in each spline, and it achieves an inference complexity of $\calO(\log_2(m_k)Lm_h)+\calO(Lm_h^2)$. 
The proposed K-DAREK architecture has $d$ MLPs with $L$ layers and $m_h$ hidden units per layer alongside one spline layer with $m_k$ knots per spline and $q$ splines (for a single-output network). 

Regardless of the number of layers, K-DAREK's error computation involves two separate knot-search stages, one for each block; this results in a total inference complexity of $\mathcal{O}(\log_2 (m_k) (d+q)) + \mathcal{O}(Lm_h^2) +\calO(qm_k)$, where the first term captures the two knot-search stages, and the second is the shared MLP dense-computation cost. The third term is the cost of evaluating the spline polynomials.
For the cost of each step in K-DAREK, please refer to  Algorithm~\ref{alg:K-DAREK}.
The computational cost is compared between different methods in  Fig.~\ref{fig:Err-comparison}~\textbf{(a)}.

\section{EXPERIMENTS AND RESULTS}
\label{sec:experiment}
We evaluated \ac{K-DAREK}~\footnote{The splines in the spline block are adopted from the KAN implementation~\cite{liu2025kan} with learnable coefficients. 
Unlike KAN, which places knots on a uniform grid, K-DAREK selects knot locations from the training data (see Knot Selection in subsection~\ref{sec:error-analysis}-1).
KAN also uses \textit{extended knots}, placing $k$ extra knots before and after the spline domain to improve boundary behavior at the domain edges. Because the true function is unknown outside the training domain, K-DAREK uses extended knots for prediction, but excludes them from the core error bound analysis. 
Their effect on model performance is evaluated in the function approximation experiment.
} against different models across multiple tasks, including function approximation, real-world regression datasets~\cite{frank2010uci}, feature collapse, overgeneralization in regions with missing data, face bounding-box prediction, and multi-agent safe control. 
We compare against \acf{GP}-based models with an \ac{RBF} kernel and learnable hyperparameters, including the length scale and variance amplitude.
These models include an exact \ac{GP} (GP)~\cite{rasmussen2006gaussian} and approximate \ac{GP} (APXGP) with learnable inducing point locations~\cite{hensman2015scalable}.
We also compare against \acl{DUE} (DUE)~\cite{van2021feature}, and \acl{SNGP} (SNGP)~\cite{liu2020simple}.
For DUE and SNGP, we use a four-layer MLP with 128 hidden units~\cite{van2021feature} as the feature extractor and estimate the corresponding posteriors using Monte Carlo (MC) sampling with 64 posterior samples.
We further compare against ensembles~\cite{lakshminarayanan2017simple} of \ac{KAN} networks, including one-layer (ENS1) and two-layer (ENS2) models.
We use $3\sigma$ intervals as uncertainty bounds for probabilistic models (GP, APXGP, DUE, SNGP, ENS1, and ENS2).
Additionally, we consider Lipschitz-based approaches DAREK and K-DAREK.
For DAREK models built upon \ac{KAN} (all-spline networks), we use DK1, which has one spline layer, and DK2, which has two stacked spline layers.
We consider two variants of our proposed model: two-layer K-DAREK (KDK2), which consists of a one-layer SNR-MLP followed by a one-layer spline network, and the three-layer K-DAREK (KDK3), which consists of a two-layer SNR-MLP followed by a one-layer spline network.
All spline layers have B-splines of degree $k=3$ (cubic B-splines). 
The Lipschitz constants for the datasets are computed numerically from the training data.

We use several metrics, including the mean squared error (MSE) is defined as $\textit{MSE} = \sum_{i=1}^n |\hat{f}(\bfx_i)-f(\bfx_i)|^2 / n$, and the root mean squared error (RMSE) is given by $\text{RMSE} = \sqrt{\sum_{i=1}^n |\hat{f}(\bfx_i)-f(\bfx_i)|^2 / n}$, where $n$ is the number of test points.
We also use the violation rate, defined as the percentage of test points where the true value lies outside the predicted error bounds.
In addition, we use the intersection over union (IoU), defined as $\text{IoU} = |\calA \cap \calB| / |\calA \cup \calB|$, where $\calA$ and $\calB$ denote the predicted and ground-truth regions, respectively~\cite{everingham2010pascal}.
We also use the sampled distance-awareness (SDA) metric defined in equation~\eqref{eq:SDA}.

\textbf{Function approximation:} 
The goal of this experiment is to learn the function $f(x)=10\cos(x)$ over the range $[-2\pi,2\pi]$ using 50 training data points. 
Both DK2 and KDK2 models are two-layer models with 5 hidden units. 
For ENS2, we use ten \ac{KAN} models with the same architecture as DK2.
All spline-based models use 9 knots and are trained for 500 epochs with a learning rate of 0.1, decayed by a factor of 0.9 every 50 epochs. 
In this experiment, DUE and SNGP use 2 layers and 20 features. \Ac{DUE} uses 9 inducing points, and \ac{SNGP} uses 9 random Fourier kernels (same as the number of knots in DK2 and KDK2). To ensure a fair comparison, we used the same input knots for the first layer in DK2, KDK2, and ENS2. The GP is trained on 9 data points selected using K-means from the training dataset. We trained DUE and SNGP for 5000 steps at a learning rate of 0.01 to ensure optimal performance, as they are very sensitive to this rate. Each experiment was repeated 10 times to compute error bars. \\
Fig.~\ref{fig:Err-comparison}~\textbf{(a)} compares the error estimation of DK2 and KDK2. The results suggest that KDK2 achieves a more generalized fit, reducing error at knots. While the interpolation error may increase slightly due to the inner block's linear approximation (rather than higher-order polynomials), the overall fit is more robust. 
In DAREK, the lack of architectural constraints can lead to erratic behavior, and the error bounds from earlier layers propagate through the model, making their control crucial. The Lipschitz constant helps justify and manage this propagation, and distributing this effect effectively is essential [cf. Remark~\ref{rem:lip-div}]. 
Fig.~\ref{fig:Err-comparison}~\textbf{(b)} compares the uncertainty bounds produced by ENS2 and \acp{GP}. 
Fig.~\ref{fig:Err-comparison}~\textbf{(c)} compares the uncertainty bounds produced by DUE and SNGP. 
Also, since DUE and SNGP use approximate GPs, their posteriors do not match the exact GP posterior. 
Table~\ref{tbl:CosineExp_combined}(a) shows that KDK2 achieves a \textit{zero} violation rate, defined as the percentage of test points where the true value lies outside the predicted error bounds, while using the fewest learnable parameters.
The ``Size'' column shows the total number of learnable parameters. 
{
The CROWN comparison shown in Fig.~\ref{fig:Err-comparison}~\textbf{(d)} is discussed further in next experiment.}

Fig.~\ref{fig:Err-comparison}~\textbf{(e)} illustrates the process time complexity of DK2 and KDK2 compared to GP, ENS2, DUE, and SNGP. 
Process time measures the total CPU work performed by the process (summed across cores).
{For this plot,} we train all the models for 30 epochs across all experiments.
The results show that KDK2 is slightly more efficient than DK2 across all dataset sizes and significantly outperforms both \ac{GP} and ENS2 for datasets larger than 300 samples. 
DUE requires higher training time and lower inference time than KDK2. 
SNGP has the lowest training and inference time among all tested models.
These results are consistent with the algorithmic complexity analysis provided in subsection~\ref{sec:algorthimic_complexity}. 

\begin{table}[!ht]
    \centering
    \caption{Error estimator comparison 
    for the function approximation experiment, with and without extended knots. \Ac{K-DAREK} achieves the lowest violation with fewer learnable parameters. The exact GP attains the lowest MSE. 
    }    
    \label{tbl:CosineExp_combined}
    \rowcolors{2}{rowgray}{white} 
    {
    \begin{tabular}{lcccccc} 
        \arrayrulecolor{headerblue}\specialrule{\heavyrulewidth}{0pt}{0pt}
        \rowcolor{headerblue}   
        Model & MSE loss & Violations(\%) & Size\\         
        
        \midrule
            \multicolumn{4}{l}{a) without extended knots}\\
            DUE     & 0.546 $\pm$ 0.075 &  \textbf{0.000 $\pm$ 0.000} & 734\\
            SNGP    & 0.396 $\pm$ 0.115 &  1.260 $\pm$ 2.179 & 726\\
            ENS2    & 1.658 $\pm$ 0.655 &  4.310 $\pm$ 6.231 & 700\\
            GP      & \textbf{0.168 $\pm$ 0.024} &  \textbf{0.000 $\pm$ 0.000} & N.A.\\
            DK2     & 0.789 $\pm$ 0.349 &  8.480 $\pm$ 4.778 & 70\\
            KDK2    & 2.151 $\pm$ 0.986 &  0.270 $\pm$ 0.359 & \textbf{45}\\
        \midrule
            \multicolumn{4}{l}{b) with extended knots}\\
            ENS2 & 0.167 $\pm$ 0.128 &  0.360 $\pm$ 0.424 & 1320\\
            DK2  & 0.102 $\pm$ 0.062 &  1.380 $\pm$ 1.992 & 132\\
            KDK2 & \textbf{0.060 $\pm$ 0.008} &  \textbf{0.000 $\pm$ 0.000} & \textbf{76}\\
        \bottomrule
    \end{tabular}
    }
    \vspace{-2mm}
\end{table}

We repeat the same experiment using \textbf{extended knots} for spline-based models.
For DK2, MSE drops from 0.789 to 0.102, and for KDK2, from 2.151 to 0.060.
KDK2 with extended knots achieves the lowest MSE and zero coverage violation while using fewer learnable parameters~(Table~\ref{tbl:CosineExp_combined} (b)).
These improvements come at the cost of additional learnable parameters (size increased from 45 to 76).
This indicates that extended knots improve smoothness, stability, and predictive accuracy. 

\textbf{Adapted-CROWN comparison experiment.}
To empirically ground the distinction between
K-DAREK's error bound and system verification methods, we compare K-DAREK against forward CROWN (implemented using \texttt{auto\_LiRPA})~\cite{xu2020automatic} on the same cosine function approximation task used in the Function Approximation experiment above, $f(x) = 10\cos(x)$ over $[-2\pi, 2\pi]$ with 50 training samples. 
CROWN certifies a fixed, already-trained network, so we used the same trained KDK2 model in both roles: as the model whose error K-DAREK bounds, and as the network CROWN certifies.
We constructed and implemented a tight affine (CROWN-style) bound using the same training knots of KDK2 as CROWN's perturbation anchors, so both methods are anchored at identical reference points. 
CROWN bounds the trained network $\hat f$: given a fixed input-perturbation set, it computes an affine relaxation from $\hat f$'s known architecture and weights, with no reference to the unknown target $f$.
K-DAREK's bound~\eqref{eq:K-DAREK}, by contrast, bounds the deviation $|f(x^*)-\hat f(x^*)|$ at a single test point $x^*$, under the assumption that $f$ is Lipschitz. 
These are bounds on different quantities — one on $\hat f$'s sensitivity over a region, the other on $\hat f$'s deviation from $f$ at a point.
The comparison here is meant to illustrate the two methods' differing design objectives, not to suggest either dominates the other.\\
We partition the input domain into intervals defined by consecutive knots. Within each intervals, splines of KDK2 are polynomial functions, $\hat f_{[j]}$. We find a tight local verification bound by solving the following optimization problem for each polynomial in every spline:
\begin{align}
    \argmin_{\alpha,\beta} &\int_{\tau_j}^{\tau_{j+1}} \big(\alpha x + \beta - \hat{f}_{[j]}(x)\big) \, dx \notag\\
    \text{s.t.} \quad & \alpha x + \beta \ge \hat{f}_{[j]}(x),
    \label{eq:crown-spline}
\end{align}
with the lower bound obtained from the mirrored problem (inequality reversed).
We solve~\eqref{eq:crown-spline} numerically via sequential least-squares quadratic programming (SLSQP); this is a new numerical procedure introduced for this comparison and is not part of DAREK's original formulation. The affine constraints are checked on a fine grid and re-certified on a finer grid, so the reported bounds are numerically validated under the adopted discretization rather than exact continuous-domain certificates. We then propagate the bounds through the entire model using \texttt{auto\_LiRPA}.\\
Fig.~\ref{fig:Err-comparison}~\textbf{(d)} illustrates the resulting bounds, together with a process-time comparison (Fig.~\ref{fig:Err-comparison}~\textbf{(e)}).
Computing the adapted-CROWN bound is substantially more expensive at query time than K-DAREK, inference takes roughly four times as long at $n=1000$ (1.10 s vs. 0.26 s), though this may reflect our unoptimized implementation of the adapted bound rather than an inherent limitation of the approach. 
Fitting the per-interval linear relaxations is a one-time cost, computed once and reused across all test queries, rather than repeated for every prediction.
The CROWN bound is tight immediately at the anchor knots. Because it is built from an independent linear relaxation fit to each knot-to-knot interval, the bound necessarily widens away from the interval's fitted point and is discontinuous across interval boundaries, which is a general property of local piecewise-linear relaxation methods, not specific to this implementation.
In the outermost intervals, bounded by padding knots outside the training domain, the bound widens substantially, since these regions extrapolate beyond where the underlying spline was fit to data.
K-DAREK's error bound (KDK2), in contrast, remains smooth and continuous throughout, widening gracefully with distance from the nearest knot.
This difference reflects the two methods' differing design objectives rather than a shortcoming of either approach.\\
Adapting CROWN itself to directly bound $|f(\bfx)-f^(\bfx)|$ is left for future work; one possible direction is incorporating training-data proximity into CROWN's relaxation, analogous to how K-DAREK's bound depends on distance to the nearest knot.

\textbf{Fixed budget experiment:}
In this experiment, we isolate the effect of architecture from model capacity by fixing the parameter count, training data, and the optimization budget across all methods.
We constrain the number of learnable parameters to about 200 for \ac{NN} models~\footnote{Table~\ref{tbl:fixed_params_budget}'s Size column reports the exact parameter count for each model. Learnable parameter counts follow directly from each architecture's discrete design choices, such as layers, hidden units, knots, and inducing points, none of which can be tuned continuously to hit exactly 200. Within this constraint, architecture choices for each model family were made to preserve a representative, non-degenerate instance of that method, rather than minimizing complexity at the expense of a baseline's typical operating regime. },  
and use a fixed training and test set of 1000 samples uniformly drawn from the 1D function $\cos(x)$ over $x \in [-2 \pi, 2 \pi]$. All models are trained for 1000 epochs with a learning rate of 0.01, which decays by a factor of 0.95 every 100 epochs.
Within this budget, we chose the architectures as:
\begin{itemize}
    \item DK2:  Two layers, 5 hidden units, 12 knots, and cubic splines with extended knots.
    \item KDK2: One dense layer and one spline layer with 10 hidden units, 10 knots, and cubic splines with extended knots.
    \item GP: \Ac{RBF}  kernel and trained on all training samples to optimize kernel hyperparameters.
    \item APXGP: \Ac{RBF} kernel and 13 inducing points.
    \item \ac{DUE}:   One residual layer, 8 hidden units, 7 inducing points, and an \ac{RBF} kernel.
    \item SNGP:  One residual layer, 10 hidden units, 25 GP features, and 20 random features.
    \item ENS2: Three KAN models, each with two spline layers, 3 hidden units, 4 knots, and cubic splines with extended knots.
\end{itemize}

We report the MSE, violation rate, size, average bound width (Avr. $u_f$), memory usage, training process time, and inference process time in Table~\ref{tbl:fixed_params_budget}. 
Each experiment is repeated 10 times, and the mean and standard deviation across runs are computed. 
ENS2 incurs the highest memory usage (13.36 KB) and the highest MSE among all models because each KAN model in the ensemble is constrained to a small parameter budget, limiting the ensemble's overall expressiveness.
The exact GP attains the lowest MSE (0.002) {and the tightest bound on average (0.02)} among the baselines but exhibits the highest violation rate (36\%) because its $3\sigma$ credible interval relies on a posterior variance that collapses as the sample density increases, and it also incurs the highest inference time (912.39 ms).
SNGP is the fastest in both training and inference times, but it requires large memory and yields high MSE and violation rates. 
KDK2 achieves the lowest MSE (0.001) and zero violations {in the evaluated test samples with a wider average bound (2.07)}. 
DK2 and KDK2 are best in training time after SNGP; however, they incur higher inference time than the probabilistic baselines with fixed budget.
The source of this overhead is architectural: unlike SNGP/DUE's single deterministic forward pass, K-DAREK additionally performs a per-feature knot search and local Newton polynomial fit to evaluate the error bound at each test point (Algorithm~\ref{alg:K-DAREK}, lines 7–8), which our current implementation executes sequentially across splines. This
can be reduced by parallel processing to 17.28 $\pm$ 1.76 ms and 22.93 $\pm$ 1.24 ms for inference time of DK2 and KDK2, respectively.

\begin{table}[!ht]
    \centering
    \caption{Fixed-budget experiment on $\cos(x)$ over $[-2\pi, 2\pi]$ with 1000 training samples and ${\sim}200$ learnable parameters per model. Process time is the total CPU work performed (summed across cores) and averaged over 10 runs after 3 warm-up passes. 
    Memory is the amount of memory required by the model's parameters. 
    K-DAREK achieves the lowest MSE and zero violations, while ranking among the fastest in training. We highlight the \textbf{best in bold}, the \sndbest{second-best in gray}, and the \red{worst in red}.}
    \label{tbl:fixed_params_budget}
    \resizebox{0.48\textwidth}{!}{%
    \rowcolors{2}{rowgray}{white}  
    \begin{tabular}{lccccccc}
        \arrayrulecolor{headerblue}\specialrule{\heavyrulewidth}{0pt}{0pt}
        \rowcolor{headerblue}     
        Model   & MSE  & Violation         &Size      & Avr. $u_f$         & Memory (KB)  & Train Time (S)    & Inf. Time (ms)       \\        
        \midrule               
        APXGP   & 0.092          & \textbf{0.000}  & 200  & 0.23    &\sndbest{0.84}& 136.573 $\pm$ 3.278      &         72.45 $\pm$   6.06  \\
        GP      & \sndbest{0.002}& \red{0.364}     & \textbf{5} & \textbf{0.02} & \textbf{0.07}& 115.021 $\pm$ 2.053 &   \red{912.39 $\pm$ 104.57} \\
        DUE     & 0.036          & \textbf{0.000}  & 204  & 1.52      & 0.93    & \red{185.972 $\pm$ 3.909}     &     \sndbest{63.34 $\pm$   6.63}  \\
        SNGP    & 0.019          & 0.037           & 201  & 0.41      & 7.05  & \textbf{44.150 $\pm$ 0.736}     &\textbf{ 10.73 $\pm$   2.52} \\
        ENS2     & \red{0.099}    & 0.028           & 216  & 0.89      &\red{13.36}   & 129.783 $\pm$ 1.297      &         68.14 $\pm$   7.25  \\
        DK2   & 0.009          &\sndbest{0.001}  & 200   & 4.09     & 5.39     & \sndbest{79.809 $\pm$ 0.906} & 144.02 $\pm$ 12.86  \\
        KDK2  & \textbf{0.001} & \textbf{0.000}  & 200   & 2.07     & 5.53         &   91.624 $\pm$ 3.009 & 220.15 $\pm$ 13.12  \\
        \bottomrule
    \end{tabular}
    }
\end{table}
{
\textbf{Ablation experiment:}
We conduct an ablation study on the same cosine function task used in the Function Approximation ($f(x) = 10 \cos(x)$, 50 training points), repeated over 10 seeded trials, and report MSE, violation rate, and average bound width.
We compare the full K-DAREK model (KDK2 with 5 hidden units, 9 knots, and cubic spline) against ablated variants that replace the MLP block with an additional spline layer (yielding an all-spline architecture), remove spectral normalization, vary the knot-selection strategy (random, Latin hypercube sampling, Chebyshev points, weighted k-means, or a fixed, equally spaced knots, in place of the plain k-means used elsewhere in this paper), and vary spline capacity (order $k=1$, or number of knots $m_k=5$ or $m_k=15$, in place of the default cubic splines, k=3, with $m_k=9$).

The results are reported in Table~\ref{tbl:ablation}. 
Replacing the MLP block with an additional spline layer increases MSE (0.108 vs. 0.060) and introduces violations (1.680\%) where the full model has none, supporting the MLP block's role in producing a more stable fit. 
Removing spectral normalization leaves MSE and violation rate largely unchanged, but the average bound width grows by roughly two orders of magnitude (2379.294 vs. 10.122), showing that without spectral normalization the MLP block's Lipschitz constant is effectively uncontrolled, producing a technically valid but highly conservative bound. 
Among knot-selection strategies, the fixed, equally spaced grid achieves the lowest MSE and zero violations in the evaluated test samples, and weighted k-means performs comparably to the plain k-means used elsewhere in the paper (0.059 vs. 0.060 MSE, both zero violations).
Random knot selection and Chebyshev points perform worse than k-means strategies, and 5 knots is too few to constrain the bound tightly, producing the widest average bound width among all variants.
The model with 15 knots achieves a low MSE, and its average bound width is smaller because the distance to the worst-case test point decreases as the number of knots increases.
}
\begin{table}[!ht]
    \centering
    {
    \caption{Ablation study (mean $\pm$ std over $N=10$ trials) isolating the effects of the MLP block, spectral normalization, knot selection strategy, and spline order/knots size ($m_k$).
    }
    \label{tbl:ablation}
    }
    \resizebox{0.48\textwidth}{!}{%
    \rowcolors{2}{rowgray}{white}
    {
    \begin{tabular}{lccc}
        \arrayrulecolor{headerblue}\specialrule{\heavyrulewidth}{0pt}{0pt}
        \rowcolor{headerblue}
        Arm & MSE & Violation (\%) & Avr. $u_f$ \\ % & Emp. Lipschitz & Curvature \\
        \midrule
        K-DAREK (KDK2)                  & 0.060 $\pm$ 0.008 & 0.000 $\pm$ 0.000  & 10.122 $\pm$ 1.525     \\ % & 10.136 $\pm$ 0.070  & 6.224 $\pm$ 0.067  \\
        No MLP block (all spline)       & 0.108 $\pm$ 0.062 & 1.680 $\pm$ 1.968  & 5.399 $\pm$ 5.606      \\ % & 19.210 $\pm$ 15.732 & 11.679 $\pm$ 8.628 \\
        No spectral norm                & 0.064 $\pm$ 0.016 & 0.000 $\pm$ 0.000  & 2379.294 $\pm$ 3116.110\\ % & 10.132 $\pm$ 0.085 & 6.224 $\pm$ 0.065 \\
        Knots: random                   & 0.336 $\pm$ 0.308 & 0.270 $\pm$ 0.447  & 94.485 $\pm$ 78.996    \\ % & 10.972 $\pm$ 0.832  & 6.437 $\pm$ 0.472 \\
        Knots: LHS                      & 0.164 $\pm$ 0.046 & 0.000 $\pm$ 0.000  & 16.186 $\pm$ 7.727     \\ % & 10.595 $\pm$ 0.399  & 6.242 $\pm$ 0.158 \\
        Knots: Chebyshev            & 0.199 $\pm$ 0.020 & 0.000 $\pm$ 0.000  & 15.180 $\pm$ 2.247 \\
        Knots: weighted k-means     & 0.059 $\pm$ 0.015 & 0.000 $\pm$ 0.000  & 11.077 $\pm$ 2.438 \\
        Knots: fixed (equally spaced)     & 0.032 $\pm$ 0.008 & 0.000 $\pm$ 0.000  & 27.164 $\pm$ 12.930    \\ 
        Spline $k=1$                    & 0.631 $\pm$ 0.164 & 0.050 $\pm$ 0.071  & 6.977 $\pm$ 0.240      \\ 
        Spline knots $m_k=5$ (small)      & 0.549 $\pm$ 0.259 & 0.010 $\pm$ 0.032  & 209.974 $\pm$ 56.002 \\
        Spline knots $m_k=15$ (large)     & 0.053 $\pm$ 0.006 & 0.070 $\pm$ 0.082  & 2.614 $\pm$ 0.529  \\
        \bottomrule
    \end{tabular}
    }
    }
\end{table}

\textbf{Sensitivity to Lipschitz misestimation}: 
We evaluate how the violation rate and average bound width respond to underestimating the Lipschitz constant. Using the same cosine function task with $5\%$ output noise (normal distribution) added to the training and test data, we scale the true Lipschitz constant $\calL_f$ by a factor ranging from 0.1 to 1.0 and recompute the resulting error bound. 
We train KDK2, a similar architecture introduced in Function approximation experiments and trained for 500
epochs with a learning rate of 0.1, decayed by a factor of 0.9 every 50 epochs, with the corresponding Lipschitz constant reported. Lipschitz constant of the noiseless true function is 10.
Table~\ref{tbl:Lipschitz_sensitivity} reports the results. RMSE remains essentially unchanged across the sweep (0.5815–0.5832). Violation rate rises sharply as the Lipschitz constant is underestimated, from 1.5\% at the full estimated value to 44.5\% at 10\% of that value, while the average bound width shrinks correspondingly (8.976 to 0.518). This confirms the mechanism described in Remark~\ref{rem:est-lip}: underestimating $\calL_f$ produces a tighter but less reliable bound, with violation rate increasing as the degree of underestimation grows. We note that, even with the fully estimated Lipschitz constant, this experiment shows that the small residual violation rate (1.6\%) is due to the added output noise rather than a limitation of the Lipschitz estimate itself.

\begin{table}[!ht]
    \centering
    
    {\caption{Sensitivity of K-DAREK's error bound to underestimating the Lipschitz constant $\calL_f$, on the cosine function task with 5\% output noise. $\calL\%$ is the fraction of the numerically estimated $\calL_f$ used to compute the bound.}
    \label{tbl:Lipschitz_sensitivity}}
    {
    \begin{tabular}{lccc}
        \arrayrulecolor{headerblue}\specialrule{\heavyrulewidth}{0pt}{0pt}
        \rowcolor{headerblue}
        $\calL\%$ & RMSE & Violation (\%) & Avr. $u_f$ \\
        \midrule
        0.10 & 0.5815 & 44.5 & 0.518 \\
        % 0.25 & 0.5830 & 23.1 & 0.940 \\
        0.50 & 0.5819 & 6.6  & 2.321 \\
        % 0.75 & 0.5830 & 2.6  & 6.043 \\
        % 0.90 & 0.5822 & 2.0  & 7.926 \\
        1.00 & 0.5832 & 1.5  & 8.976 \\
        \bottomrule
    \end{tabular}}
\end{table}

\textbf{Real data regression experiment:}
\label{sec:real-data}
We evaluate model performance on six real-world regression datasets from the UCI repository~\cite{frank2010uci}: 
% Concrete, Airfoil Self-Noise, Combined Cycle Power Plant, Red Wine, White Wine, and Real Estate Valuation.
Concrete Compressive Strength ($n= 1030$, 7 continuous features), which predicts the compressive strength of concrete from its ingredients and age; 
Airfoil Self-Noise ($n= 1503$, 3 continuous features), which predicts aerodynamic noise generated by airfoils from operating conditions; 
Combined Cycle Power Plant ($n= 9568$, 4 continuous features), which predicts the net electrical power output of a power plant from environmental variables;
Red and White Wine Quality (Red Wine has $n= 1599$, 11 continuous features and White Wine has $n= 4898$, 11 continuous features), which predict wine quality scores from physicochemical measurements; and Real Estate Valuation ($n= 414$, 6 continuous features), which predicts residential property values from location- and property-related attributes. We report only the number of continuous input features, as categorical features are excluded from our experiments.
For results in Table~\ref{tbl:real_data_2features}, models are trained using only the first two input features, whereas Table~\ref{tbl:real_data_allfeatures} reports results using all \textit{continuous} features. 
Each experiment is repeated 20 times, and we report the mean and standard deviation of RMSE and violation rate across runs for all datasets. 

Training is performed on the standardized dataset with the Adam optimizer and a decaying learning rate.
For the Power Plant and White Wine datasets, we use a subset of 2000 samples to make running the GP computationally feasible.
All models have five hidden units, and ENS2 consists of five models.
The \ac{MLP} model has two layers with 64 hidden units, and the \ac{KAN} model has two layers with 5 hidden units.
DUE uses 20 inducing points, and SNGP uses 1024 random Fourier kernels.
MLP, KAN, GP, DK1, DK2, KDK2, KDK3, and ENS2 are trained for 1000 epochs using the Adam optimizer, starting with a learning rate of 0.1 and applying a decay rate of 0.9 every 100 epochs. 
DUE and SNGP were trained for 5000 steps with a learning rate of 0.01.

In most experiments, the proposed models (KDK2 and KDK3) achieve the best or the second-best performance in terms of RMSE or violation rate.
The results demonstrate that 
% Lipschitz-certified uncertainty bounds provide efficiently computable theoretical guarantees and 
{Lipschitz-based error bounds}
achieve stronger empirical coverage than \ac{DUE}, \ac{SNGP}, and ENS2.

We note that a low violation rate alone does not establish that a bound is tight, since a sufficiently conservative bound trivially achieves zero violations. Remark~\ref{rem:tight-condition}  gives the condition under which our interpolation bound is tight (equality in~\eqref{eq:interpolation-error}).

\begin{table*}
    \centering
    {\tiny
    \caption{
		Results of test errors and percentage of violations on real datasets \textit{over the first two input features}. \textbf{Bold} and \sndbest{gray bold italic} values indicate the best and second-best models, respectively. The proposed method achieves the fewest violations across datasets; RMSE is competitive with, but not uniformly lower than, the best baseline for each dataset.
        }\label{tbl:real_data_2features}
        }		
	\centering
        \setlength\extrarowheight{2pt}  
    \resizebox{\textwidth}{!}{%
	\begin{tabular}{lTTTTTTVVVVVV}
        \cline{1-13}
        \multicolumn{1}{c}{} & \multicolumn{6}{c}{RMSE} & \multicolumn{6}{c}{Violations (\%)} \\
		\cline{2-6}
        \cline{7-13}
		Model & Concrete & Airfoil Self-Noise & Power Plant & Red Wine & White Wine & Real Estate & Concrete & Airfoil Self-Noise & Power Plant & Red Wine & White Wine & Real Estate\\
                      
		\hline
            MLP& 0.79 $\pm$ 0.05 & 0.79 $\pm$ 0.03 & \textbf{0.27 $\pm$ 0.01} & 0.90 $\pm$ 0.02 & 0.97 $\pm$ 0.04 & 0.95 $\pm$ 0.12 & - & - & - & - & - & - \\
            KAN& 0.84 $\pm$ 0.05 & 0.85 $\pm$ 0.07 & 0.30 $\pm$ 0.04 & 0.93 $\pm$ 0.04 & 0.98 $\pm$ 0.05 & 1.04 $\pm$ 0.14 & - & - & - & - & - & - \\
            GP & \sndbest{0.79 $\pm$ 0.04} & \sndbest{0.78 $\pm$ 0.06} & \textbf{0.27 $\pm$ 0.01} & \textbf{0.88 $\pm$ 0.03} & \textbf{0.94 $\pm$ 0.05} & \sndbest{0.90 $\pm$ 0.12} & \sndbest{0.05 $\pm$ 0.15} & \sndbest{0.53 $\pm$ 0.73} & \sndbest{0.87 $\pm$ 0.26} & 0.69 $\pm$ 0.27 & \sndbest{0.60 $\pm$ 0.39} & 0.84 $\pm$ 0.94 \\            
            DUE& 0.98 $\pm$ 0.06 & 0.98 $\pm$ 0.05 & 0.85 $\pm$ 0.29 & 0.96 $\pm$ 0.03 & 0.97 $\pm$ 0.05 & 1.00 $\pm$ 0.13 & 33.16 $\pm$ 2.84 & 32.36 $\pm$ 3.28 & 37.33 $\pm$ 4.54 & 15.69 $\pm$ 1.14 & 27.10 $\pm$ 2.57 & 32.41 $\pm$ 5.74 \\
            SNGP& 0.99 $\pm$ 0.07 & 1.00 $\pm$ 0.04 & 1.00 $\pm$ 0.03 & 0.96 $\pm$ 0.03 & 0.98 $\pm$ 0.05 & 1.00 $\pm$ 0.13 & 76.89 $\pm$ 5.28 & 79.77 $\pm$ 6.01 & 91.78 $\pm$ 1.83 & 94.50 $\pm$ 10.82 & 63.82 $\pm$ 15.19 & 62.77 $\pm$ 14.30 \\
            DK1& 0.80 $\pm$ 0.06 & 0.85 $\pm$ 0.03 & \textbf{0.27 $\pm$ 0.01} & 0.89 $\pm$ 0.03 & 1.03 $\pm$ 0.19 & \textbf{0.90 $\pm$ 0.11} & \textbf{0.00 $\pm$ 0.00} & 1.03 $\pm$ 0.69 & \textbf{0.00 $\pm$ 0.00} & \sndbest{0.38 $\pm$ 0.27} & 2.12 $\pm$ 0.99 & \sndbest{0.24 $\pm$ 0.48} \\
            DK2& 0.85 $\pm$ 0.11 & 0.86 $\pm$ 0.34 & 0.30 $\pm$ 0.05 & 0.96 $\pm$ 0.04 & 1.00 $\pm$ 0.06 & 1.16 $\pm$ 0.11 & \textbf{0.00 $\pm$ 0.00} & \textbf{0.00 $\pm$ 0.00} & \textbf{0.00 $\pm$ 0.00} & \textbf{0.00 $\pm$ 0.00} & \textbf{0.00 $\pm$ 0.00} & \textbf{0.00 $\pm$ 0.00} \\
            KDK2& \textbf{0.78 $\pm$ 0.05} & 0.82 $\pm$ 0.03 & \textbf{0.27 $\pm$ 0.01} & \sndbest{0.89 $\pm$ 0.02} & 0.97 $\pm$ 0.04 & 0.98 $\pm$ 0.12 & \textbf{0.00 $\pm$ 0.00} & \textbf{0.00 $\pm$ 0.00} & \textbf{0.00 $\pm$ 0.00} & \textbf{0.00 $\pm$ 0.00} & \textbf{0.00 $\pm$ 0.00} & \textbf{0.00 $\pm$ 0.00} \\
            KDK3 & 0.90 $\pm$ 0.18 & 0.84 $\pm$ 0.04 & 0.47 $\pm$ 0.04 & 0.89 $\pm$ 0.03 & \sndbest{0.95 $\pm$ 0.05} & 0.94 $\pm$ 0.12 & \textbf{0.00 $\pm$ 0.00} & \textbf{0.00 $\pm$ 0.00} & \textbf{0.00 $\pm$ 0.00} & \textbf{0.00 $\pm$ 0.00} & \textbf{0.00 $\pm$ 0.00} & \textbf{0.00 $\pm$ 0.00} \\     
            ENS2& \sndbest{0.79 $\pm$ 0.04} & \textbf{0.77 $\pm$ 0.04} & \sndbest{0.29 $\pm$ 0.02} & \sndbest{0.89 $\pm$ 0.02} & 0.97 $\pm$ 0.07 & 0.97 $\pm$ 0.11 & 39.85 $\pm$ 10.02 & 30.33 $\pm$ 7.04 & 30.87 $\pm$ 11.83 & 42.78 $\pm$ 7.58 & 44.90 $\pm$ 7.40 & 28.43 $\pm$ 8.07 \\
		\hline 
	\end{tabular}}
\end{table*}

\begin{table*}
    \centering
    {\tiny
    \caption{
		Results of test errors and percentage of violations on real datasets \textit{over all continuous input features}. Bold and \sndbest{gray bold italic}  values indicate the best and second-best models, respectively. The proposed method achieves the fewest violations across datasets; RMSE is competitive with, but not uniformly lower than, the best baseline for each dataset.         
        }
		\label{tbl:real_data_allfeatures}}
	\centering
        \setlength\extrarowheight{2pt} 
    \resizebox{\textwidth}{!}{%
	\begin{tabular}{lTTTTTTVVVVVV}
        \cline{1-13}
        \multicolumn{1}{c}{} & \multicolumn{6}{c}{RMSE} & \multicolumn{6}{c}{Violations (\%)} \\
		\cline{2-7}
        \cline{8-13}
	   Model & Concrete & Airfoil Self-Noise & Power Plant & Red Wine & White Wine & Real Estate & Concrete & Airfoil Self-Noise & Power Plant & Red Wine & White Wine & Real Estate\\
                            
		\hline            
            MLP& 0.78 $\pm$ 0.06 & \sndbest{0.37 $\pm$ 0.04} & \textbf{0.25 $\pm$ 0.01} & 1.00 $\pm$ 0.10 & 0.95 $\pm$ 0.04 & 0.85 $\pm$ 0.13 & - & - & - & - & - & -  \\
            KAN& 0.94 $\pm$ 0.12 & 0.69 $\pm$ 0.10 & 0.55 $\pm$ 0.53 & 2.06 $\pm$ 1.18 & 3.55 $\pm$ 3.20 & 3.99 $\pm$ 5.84 & - & - & - & - & - & -  \\
            GP & \textbf{0.70 $\pm$ 0.05} & \textbf{0.32 $\pm$ 0.03} & \textbf{0.25 $\pm$ 0.01} & \textbf{0.74 $\pm$ 0.02} & \sndbest{0.83 $\pm$ 0.07} & \textbf{0.55 $\pm$ 0.13} & \sndbest{0.05 $\pm$ 0.15} & \sndbest{1.66 $\pm$ 0.76} & \sndbest{0.68 $\pm$ 0.20} & 0.91 $\pm$ 0.41 & \sndbest{0.70 $\pm$ 0.33} & \sndbest{1.33 $\pm$ 1.14} \\
            DUE& 0.75 $\pm$ 0.05 & 0.33 $\pm$ 0.02 & \textbf{0.24 $\pm$ 0.01} & 0.82 $\pm$ 0.04 & 0.77 $\pm$ 0.03 & 0.68 $\pm$ 0.14 & 40.24 $\pm$ 5.69 & 24.49 $\pm$ 2.00 & 31.95 $\pm$ 2.39 & 52.06 $\pm$ 2.68 & 51.28 $\pm$ 2.66 & 46.75 $\pm$ 8.85 \\
            SNGP& 0.98 $\pm$ 0.06 & 1.00 $\pm$ 0.04 & 1.00 $\pm$ 0.02 & 0.96 $\pm$ 0.03 & 0.98 $\pm$ 0.05 & 1.00 $\pm$ 0.13 & 18.30 $\pm$ 3.12 & 64.55 $\pm$ 4.61 & 58.67 $\pm$ 3.69 & 6.34 $\pm$ 0.96 & 18.05 $\pm$ 2.41 & 1.33 $\pm$ 0.84 \\
            DK1& \sndbest{0.71 $\pm$ 0.04} & 0.69 $\pm$ 0.03 & \textbf{0.25 $\pm$ 0.01} & \sndbest{0.79 $\pm$ 0.03} & \textbf{0.81 $\pm$ 0.03} & \sndbest{0.60 $\pm$ 0.11} & \sndbest{0.05 $\pm$ 0.15} & \textbf{0.00 $\pm$ 0.00} & 1.12 $\pm$ 0.70 & 25.00 $\pm$ 8.51 & 10.32 $\pm$ 3.14 & 2.05 $\pm$ 1.53 \\
            DK2& 0.84 $\pm$ 0.07 & \sndbest{0.37 $\pm$ 0.04} & \sndbest{0.26 $\pm$ 0.02} & 0.90 $\pm$ 0.03 & 0.91 $\pm$ 0.04 & 0.87 $\pm$ 0.16 & \textbf{0.00 $\pm$ 0.00} & \textbf{0.00 $\pm$ 0.00} & \textbf{0.00 $\pm$ 0.00} & \textbf{0.00 $\pm$ 0.00} & \textbf{0.00 $\pm$ 0.00} & \textbf{0.00 $\pm$ 0.00} \\
            KDK2& 0.79 $\pm$ 0.05 & 0.57 $\pm$ 0.03 & 0.27 $\pm$ 0.01 & 0.88 $\pm$ 0.05 & 0.91 $\pm$ 0.04 & 0.68 $\pm$ 0.14 & \textbf{0.00 $\pm$ 0.00} & \textbf{0.00 $\pm$ 0.00} & \textbf{0.00 $\pm$ 0.00} & \sndbest{0.75 $\pm$ 0.45} & \textbf{0.00 $\pm$ 0.00} & \textbf{0.00 $\pm$ 0.00} \\
            KDK3& 0.82 $\pm$ 0.05 & 0.81 $\pm$ 0.26 & 0.51 $\pm$ 0.02 & 0.86 $\pm$ 0.03 & 0.92 $\pm$ 0.04 & 0.72 $\pm$ 0.14 & \textbf{0.00 $\pm$ 0.00} & \textbf{0.00 $\pm$ 0.00} & \textbf{0.00 $\pm$ 0.00} & \textbf{0.00 $\pm$ 0.00} & \textbf{0.00 $\pm$ 0.00} & \textbf{0.00 $\pm$ 0.00} \\
            ENS2& 0.98 $\pm$ 0.24 & 0.55 $\pm$ 0.07 & 0.34 $\pm$ 0.07 & 2.20 $\pm$ 1.33 & 1.14 $\pm$ 0.27 & 1.21 $\pm$ 0.62 & 20.68 $\pm$ 3.81 & 9.10 $\pm$ 3.53 & 8.10 $\pm$ 4.15 & 11.41 $\pm$ 2.05 & 15.90 $\pm$ 2.71 & 4.22 $\pm$ 2.17 \\
		\hline 
	\end{tabular}}
\end{table*}

\begin{figure*}[t]
\centering
\setlength{\tabcolsep}{1pt}

\includegraphics[width=0.98\linewidth]{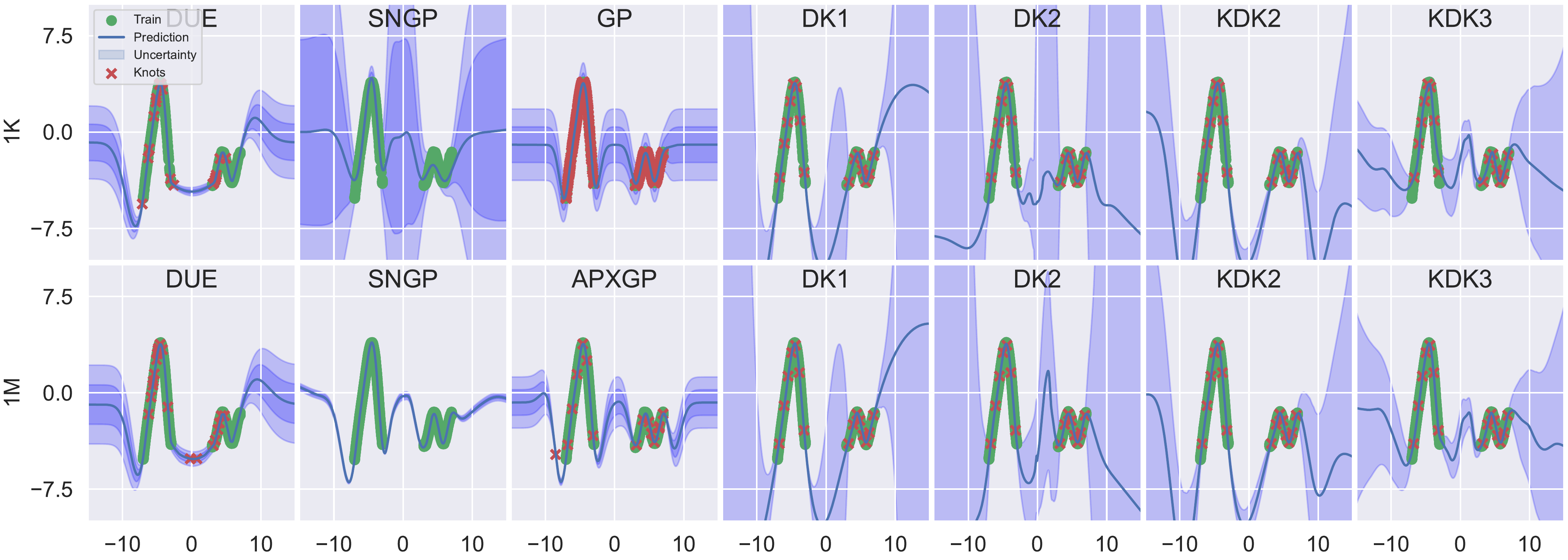}

\caption{Mean of prediction and uncertainty estimates for a 1D regression task, used to evaluate feature collapse and over-generalization across different methods at two training set sizes (1K and 1M samples). 
Red cross \red{$\mathbf{\times}$} are projected inducing points in input space for DUE, inducing points for APXGP, and knots for DK1, DK2, KDK2, and KDK3; Note that some are placed in regions with no nearby training samples.
GP is replaced by APXGP in the large dataset due to scalability constraints.
All the models except SNGP estimate a large uncertainty in the gap between observed regions. }
\label{fig:feature-collapse}
\end{figure*}

\textbf{Feature collapse experiment.}
We repeat the experiment in Fig. 1 of~\cite{van2021feature}, using the same reported setup for DUE and SNGP, to evaluate whether \ac{DAREK} and \ac{K-DAREK} suffer from feature collapse and whether they overgeneralize in uncertainty estimation. We additionally include GP and APXGP in the experiments. 
There are two datasets: a small dataset with 1k training samples and a large dataset with 1000k training samples. 
All spline-based models (DK2, DK3, KDK2, KDK3) use 15 knots per spline and 5 units for intermediate layers.
APXGP uses 15 inducing points. 
As shown in Fig.~\ref{fig:feature-collapse}, {among the compared methods}, \ac{SNGP} tends to overgeneralize {as the dataset grows, with} its uncertainty converges to zero; 
GP/APXGP, DUE, and our method (DK/KDK) all avoid this failure mode at both dataset sizes.

\textbf{Uncertainty of missing data (Lipschitz vs. Probabilistic perspective).} 
Probabilistic models estimate uncertainty based on observed data and model assumptions (such as priors); however, Lipschitz-based methods provide worst-case guarantees independent of data distribution. 
This experiment explicitly shows this difference on datasets where the ground-truth function contains localized structure that is entirely unobserved during training.  
We construct one-dimensional synthetic datasets as: 
\[f(x) = g(x) + 10\exp(-\frac{(x)^2}{8}),\]
which has a Gaussian perturbation centered in the middle of the domain, added to a base function $g(x)$.
We consider five variants of $g(x)$: zero, linear ($x/2$), quadratic ($(x/2)^2$), a sinusoid ($3\sin(x)$), and a $\tanh$ function ($5\tanh(x/2)$). For each variant,
we uniformly draw 50 samples from the interval $[-7.5,-3] \cap [3,7.5]$.
To simulate a realistic out-of-distribution (OOD) scenario, all training samples within a central interval ${x \in [-3,3]}$ are removed.
{Models are trained for 1000 epochs outside the gap and evaluated over the full domain, with the evaluation repeated across 10 seeded trials per g(x) variant.}

\begin{figure}
    \centering
    \includegraphics[width=0.98\linewidth]{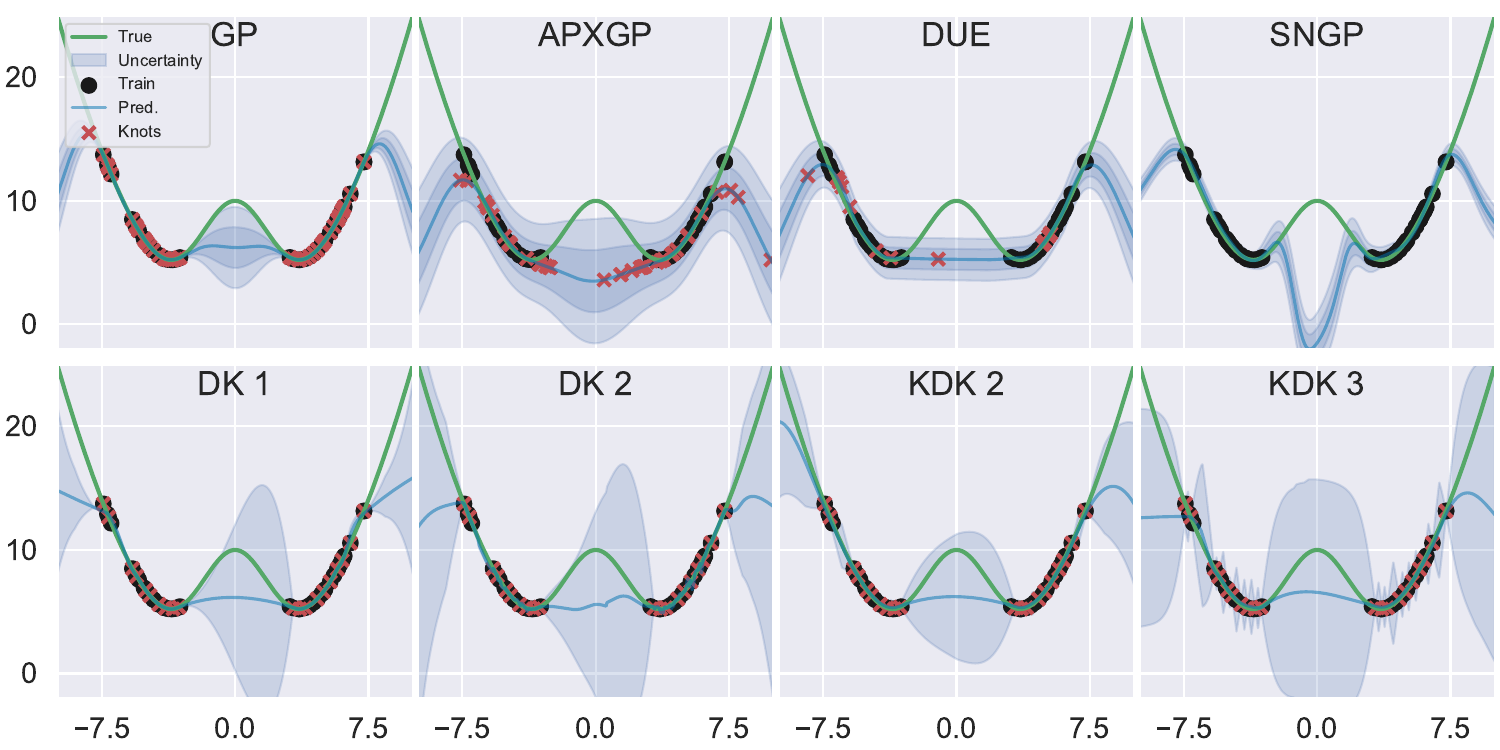}
    \caption{Uncertainty quantification of probabilistic methods versus Lipschitz-based methods in the presence of missing data.}
    \label{fig:unc-miss-data}
\end{figure}

Fig.~\ref{fig:unc-miss-data} illustrates predictive means and uncertainty estimates across the entire input domain for the quadratic case ($g(x) = (x/2)^2$).
None of the models recover the localized Gaussian perturbation, as expected given the absence of training data in that region, although predictive uncertainty increases within the gap {for most models}.
GP and APXGP expand in the missing data region, whereas DUE and SNGP do not recover comparable expansions; 
the projections of some of the inducing points for both DUE and APXGP lie in the missing area.
In contrast, \ac{DAREK} and \ac{K-DAREK} expand significantly in the missing data region, depending on the estimated Lipschitz constant and the distance to the training set.

Table~\ref{tbl:multi_g} reports gap-region ([-3,3]) violation rate and average uncertainty for GP, SNGP, DUE, and KDK2 across all $g(x)$ variants. 
KDK2 achieves zero gap-region violations across all variants. 
SNGP and DUE show substantial violations throughout (54–98\%), consistent with the overgeneralization visible in Fig.~\ref{fig:unc-miss-data}. 
GP achieves zero violations in four of the five variants, failing only on the $\tanh$ case. We note that KDK2's zero-violation coverage comes at the cost of a wider average bound than the probabilistic baselines in every variant, reflecting the same tightness-coverage tradeoff observed in the Fixed Budget experiment.

\begin{table}[!ht]
    \centering
    \caption{Gap-region ($x \in [-3,3]$) violation rate (\%) and average uncertainty width, across five base functions $g(x)$, with the bump term $10\exp(-x^2/8)$ held fixed.}
    \label{tbl:multi_g}
    \resizebox{0.48\textwidth}{!}{%
    \begin{tabular}{l|cccc|cccc|}
        \arrayrulecolor{headerblue}\specialrule{\heavyrulewidth}{0pt}{0pt}
        \rowcolor{headerblue}
        &\multicolumn{4}{|c}{Violation (\%)} & \multicolumn{4}{|c|}{Average uncertainty $u_f$} \\
        \midrule
        $g(x)$ & GP & SNGP & DUE & KDK2 & GP & SNGP & DUE & KDK2 \\
        \midrule
        zero      & 0.000 & 79.100 & 98.467 & \textbf{0.000} & 0.572 & 2.881 & 0.337 & 7.052 \\
        linear    & 0.000 & 78.533 & 97.433 & \textbf{0.000} & 0.614 & 3.205 & 0.428 & 9.172 \\
        quadratic & 0.000 & 64.100 & 54.433 & \textbf{0.000} & 2.453 & 3.297 & 2.662 & 30.704\\
        sine      & 0.000 & 72.067 & 79.900 & \textbf{0.000} & 2.287 & 3.025 & 0.697 & 36.805\\
        tanh      & 7.500 & 73.467 & 74.833 & \textbf{0.000} & 0.646 & 3.276 & 1.746 & 15.277\\
        \bottomrule
    \end{tabular}
    }
\end{table}

\textbf{High dimension distance-awareness.}
We evaluated \ac{K-DAREK} on the Celebrity Attribute face detection dataset~\cite{liu2015deep}. 
The task is to predict the bounding box of the face in the given image.
To reduce the dataset size, we randomly selected 8k images for training and 1k images for testing.
We reduced feature dimensionality to (2, 5, 10, 20, 50, 100, and 200) using two feature extraction strategies: PCA components of input images and PCA components of ResNet-18 features. 
The RMSE, IoU, and SDA are reported in Table~\ref{tbl:facerecognition}.

For the first approach, we resize the images to a fixed dimension of $224{\times}224{\times}3$ and then select the $n$ top PCA components.
For the second approach, we resize the images to a fixed dimension of $224{\times}224{\times}3$, apply the feature extractor of pretrained ResNet-18 to get $7{\times}7{\times}512$ features, and select the $n$ top PCA components.
Each of ENS1 and ENS2 has 5 KAN models.
The DK1, DK2, KDK2, KDK3, ENS1, and ENS2 models have 10 hidden units and 15 knots for each spline.
GP was trained on 15 training points selected using K-means from the training dataset, and DUE and APXGP use 15 inducing points. SNGP uses 1024 random Fourier kernels.
All models were trained with a learning rate of 0.1 and decayed by a factor of 0.9 every 200 iterations.

\begin{figure}
    \centering
    \includegraphics[width=0.9\linewidth]{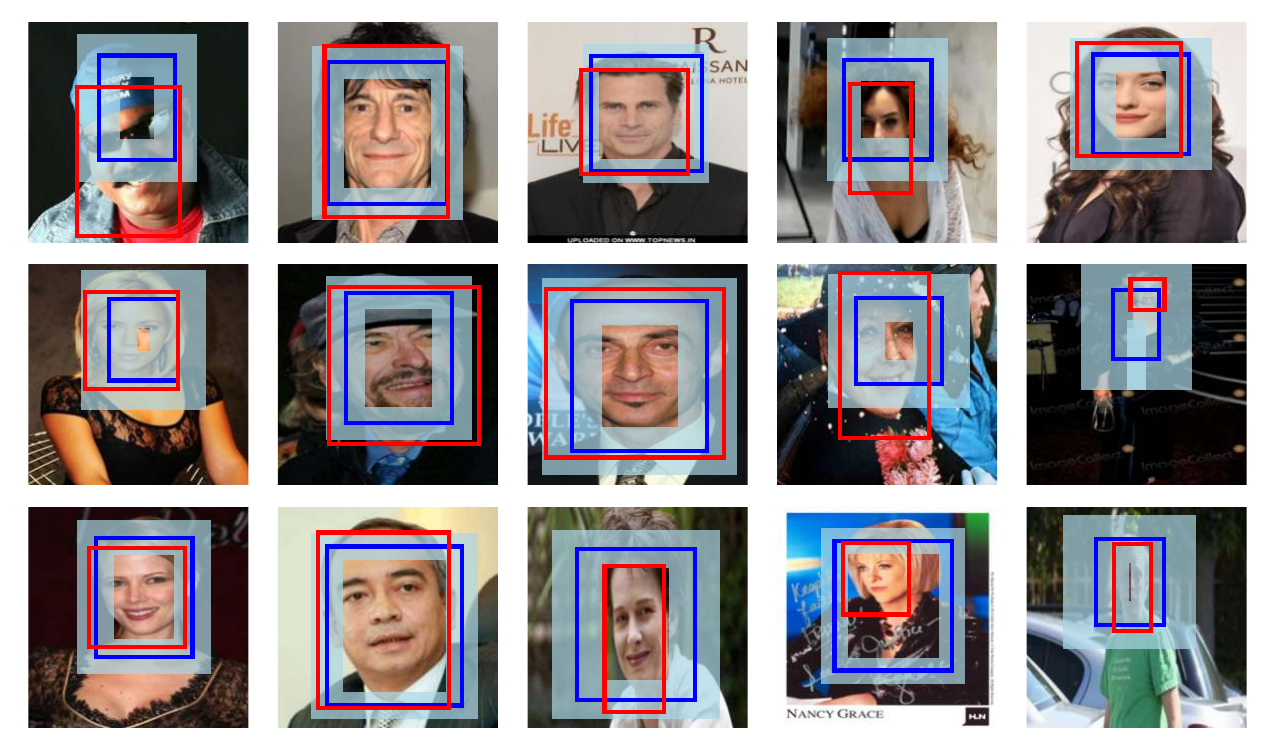}
    \caption{Face bounding-box prediction on sample test images from the CelebA dataset~\cite{liu2015deep} using K-DAREK (KDK2). \red{Red boxes} indicate ground-truth bounding boxes and \blue{blue boxes} indicate K-DAREK predictions. The \cyan{shaded blue} region indicates the error estimates in the x- and y-coordinates. Note that it uses only 100 projected dimensions out of a 25k-dimensional feature space of ResNet-18. }
    \label{fig:face_recognition}
\end{figure}

GP was trained for 2000 epochs, DUE for 5000 epochs, SNGP for 10000 epochs, and the remaining models for 1000 iterations.
RMSE and IoU remain consistent across models and features, with ResNet-18 features yielding lower RMSE and higher IoU.
DAREK's and K-DAREK's SDA~\eqref{eq:SDA} increase from approximately 60\% at two dimensions to around 95\% at 100 dimensions, outperforming ENS1 and ENS2, which do not follow a consistent pattern. 
We believe the pattern in SDA of DAREK and K-DAREK reflects a curse-of-dimensionality effect: as the input dimension grows, training data becomes sparser relative to the volume of the input space, so nearest-knot distances tend to grow with dimension.  We do not have a rigorous account of this effect and leave it to future work.
GP-based models achieve near-perfect SDA due to their kernel-based formulation.
We evaluated SNGP's SDA metric with respect to training points selected by the K-means approach, since SNGP does not have specific training representatives (such as inducing points or knots). 
Fig.~\ref{fig:face_recognition} shows a representative mix of correctly bounded and violated predictions. Violations occur where the numerically estimated Lipschitz constant $\calL_f$ underestimates the true local variation of the target function, due to the MLP block's limited capacity on the coarse, low-dimensional feature projection used here, and occasional label noise in the CelebA ground-truth bounding boxes. 

\begin{table*}[t]
\centering
\caption{Results for high-dimensional experiments across two feature extraction methods, PCA and ResNet-18. }
\label{tbl:facerecognition}
\setlength{\tabcolsep}{4pt}
\renewcommand{\arraystretch}{1.05}

\rowcolors{7}{rowgray}{white}  
\resizebox{\textwidth}{!}{%
\begin{tabular}{l|ccccccc|ccccccc|ccccccc}
\arrayrulecolor{black}\cline{1-22}
\rowcolor{headerblue}
& \multicolumn{7}{c|}{RMSE $\downarrow$} 
& \multicolumn{7}{c|}{IoU $\uparrow$} 
& \multicolumn{7}{c}{SDA $\uparrow$} \\

\arrayrulecolor{black}\cline{2-22}

\rowcolor{headerblue}
Model 
& 2 & 5 & 10 & 20 & 50 & 100 & 200
& 2 & 5 & 10 & 20 & 50 & 100 & 200
& 2 & 5 & 10 & 20 & 50 & 100 & 200 \\

\arrayrulecolor{black}\cline{1-22}
\noalign{\vspace{4pt}}
\multicolumn{22}{l}{\textbf{(a) PCA components from images}}\\
DK1
  & 0.136 & 0.133 & 0.133 & 0.133 & 0.126 & 0.129 & 0.149
  & 0.451 & 0.456 & 0.454 & 0.452 & 0.469 & 0.466 & 0.410
  & 0.676 & 0.856 & 0.947 & 0.967 & 0.974 & 0.952 & 0.969 \\
 
DK2
  & 0.137 & 0.139 & 0.134 & 0.190 & 0.123 & 0.184 & 0.135
  & 0.448 & 0.443 & 0.454 & 0.375 & 0.486 & 0.339 & 0.451
  & 0.677 & 0.857 & 0.947 & 0.972 & 0.974 & 0.950 & 0.971 \\

KDK2 
  & 0.137 & 0.133 & 0.130 & 0.130 & 0.134 & 0.135 & 0.136 
  & 0.439 & 0.450 & 0.456 & 0.464 & 0.444 & 0.446 & 0.440 
  & 0.726 & 0.843 & 0.918 & 0.929 & 0.971 & 0.945 & 0.978 \\
  
KDK3 
  & 0.137 & 0.131 & 0.130 & 0.128 & 0.130 & 0.132 & 0.133 
  & 0.447 & 0.459 & 0.464 & 0.470 & 0.462 & 0.456 & 0.454 
  & 0.692 & 0.741 & 0.862 & 0.919 & 0.917 & 0.911 & 0.932 \\

ENS1
  & 0.136 & 0.133 & 0.133 & 0.133 & 0.126 & 0.130 & 0.151
  & 0.450 & 0.457 & 0.455 & 0.451 & 0.467 & 0.464 & 0.407
  & 0.571 & 0.554 & 0.568 & 0.563 & 0.501 & 0.431 & 0.378 \\
 
ENS2
  & 0.135 & 0.131 & 0.131 & 0.128 & 0.183 & 0.265 & 0.129
  & 0.449 & 0.456 & 0.446 & 0.469 & 0.311 & 0.145 & 0.431
  & 0.527 & 0.542 & 0.545 & 0.530 & 0.588 & 0.564 & 0.525 \\

GP 
 & 0.138 & 0.149 & 0.137 & 0.161 & 0.138 & 0.137 & 0.143
 & 0.435 & 0.400 & 0.450 & 0.359 & 0.437 & 0.448 & 0.420
 & 1.000 & 1.000 & 1.000 & 1.000 & 1.000 & 1.000 & 1.000 \\
 
APXGP
  & 0.138 & 0.137 & 0.137 & 0.137 & 0.137 & 0.137 & 0.137
  & 0.427 & 0.447 & 0.447 & 0.447 & 0.447 & 0.447 & 0.447
  & 0.977 & 1.000 & 1.000 & 1.000 & 1.000 & 1.000 & 1.000 \\

DUE 
  & 0.135 & 0.130 & 0.130 & 0.119 & 0.103 & 0.104 & 0.112 
  & 0.450 & 0.470 & 0.480 & 0.523 & 0.554 & 0.556 & 0.534 
  & 0.682 & 0.930 & 0.993 & 1.000 & 1.000 & 1.000 & 1.000 \\
  
SNGP & 0.138 & 0.144 & 0.164 & 0.155 & 0.148 & 0.198 & 0.207 & 0.445 & 0.443 & 0.421 & 0.424 & 0.427 & 0.244 & 0.246 & 0.628 & 0.840 & 0.892 & 0.902 & 0.867 & 0.553 & 0.593 \\
\arrayrulecolor{black}\cline{1-22}
\noalign{\vspace{4pt}}
\multicolumn{22}{l} {\textbf{(b) PCA components from ResNet-18 features}}\\
DK1
  & 0.100 & 0.097 & 0.090 & 0.088 & 0.079 & 0.074 & 0.079
  & 0.535 & 0.538 & 0.562 & 0.557 & 0.586 & 0.606 & 0.589
  & 0.704 & 0.686 & 0.875 & 0.892 & 0.788 & 0.837 & 0.935 \\
 
DK2
  & 0.100 & 0.100 & 0.093 & 0.103 & 0.083 & 0.073 & 0.072
  & 0.533 & 0.535 & 0.553 & 0.470 & 0.573 & 0.608 & 0.626
  & 0.691 & 0.707 & 0.899 & 0.918 & 0.858 & 0.882 & 0.902 \\

KDK2 
  & 0.100 & 0.095 & 0.086 & 0.086 & 0.077 & 0.089 & 0.071 
  & 0.536 & 0.548 & 0.575 & 0.575 & 0.602 & 0.493 & 0.640 
  & 0.700 & 0.734 & 0.819 & 0.869 & 0.849 & 0.860 & 0.904 \\
KDK3 
  & 0.101 & 0.095 & 0.086 & 0.087 & 0.084 & 0.090 & 0.130 
  & 0.531 & 0.546 & 0.574 & 0.567 & 0.553 & 0.518 & 0.381 
  & 0.690 & 0.724 & 0.791 & 0.830 & 0.849 & 0.897 & 0.908 \\
  
ENS1 
  & 0.100 & 0.097 & 0.089 & 0.089 & 0.080 & 0.075 & 0.081 
  & 0.535 & 0.540 & 0.565 & 0.557 & 0.585 & 0.603 & 0.583 
  & 0.578 & 0.551 & 0.566 & 0.562 & 0.364 & 0.222 & 0.408 \\

ENS2
  & 0.100 & 0.094 & 0.087 & 0.084 & 0.174 & 0.069 & 0.065
  & 0.530 & 0.551 & 0.569 & 0.575 & 0.161 & 0.628 & 0.646
  & 0.515 & 0.533 & 0.573 & 0.562 & 0.461 & 0.550 & 0.542\\

GP
 & 0.139 & 0.140 & 0.139 & 0.140 & 0.140 & 0.137 & 0.138
 & 0.431 & 0.447 & 0.434 & 0.427 & 0.426 & 0.447 & 0.444
 & 0.542 & 1.000 & 1.000 & 1.000 & 1.000 & 1.000 & 1.000 \\

APXGP
  & 0.133 & 0.137 & 0.137 & 0.137 & 0.137 & 0.137 & 0.137
  & 0.445 & 0.447 & 0.447 & 0.447 & 0.447 & 0.447 & 0.447
  & 0.729 & 1.000 & 1.000 & 1.000 & 1.000 & 1.000 & 1.000 \\

DUE 
  & 0.099 & 0.095 & 0.087 & 0.085 & 0.076 & 0.072 & 0.068 
  & 0.537 & 0.548 & 0.581 & 0.575 & 0.608 & 0.637 & 0.650 
  & 0.722 & 0.937 & 0.990 & 0.997 & 1.000 & 1.000 & 1.000 \\

SNGP 
  & 0.102 & 0.095 & 0.107 & 0.166 & 0.176 & 0.204 & 0.245 
  & 0.528 & 0.547 & 0.525 & 0.341 & 0.305 & 0.229 & 0.138 
  & 0.591 & 0.869 & 0.832 & 0.544 & 0.560 & 0.497 & 0.512 \\

\arrayrulecolor{black}\cline{1-22}
\end{tabular}}

\end{table*}

\textbf{Multi-agent safe control:} 
Learning-based control integrates data-driven modeling techniques, such as \acp{NN}, with traditional control theory to handle complex or partially unknown dynamical systems. These methods enhance adaptability and robustness, especially in environments where analytical models are difficult to derive, or the system must respond to uncertainty~\cite{dhiman2021control,brunke2022safe,lederer2019uniform,capone2020localized,long2026sensor,wang2021learning,lindemann2021learning,fan2020bayesian,gahlawat2020l1,marvi2020safe,srinivasan2020synthesis,kim2021hamilton,sarkar2021finite,fattahi2019learning,buisson2020actively,rahimi2022robust}.

In this experiment, we evaluated both DAREK and \ac{K-DAREK} in a multi-agent safe control~\cite{zhang2025gcbf+,prorok2021beyond,borrmann2015control,beckers2021online,jing2026probabilistic,navsalkar2023data} setup adopted from~\cite{cheng2020safe}. 
Fig.~\ref{fig:Safe-Control} shows a diagram of the training and control loop beside a successful trajectory. 
 The controlled agent (blue car) must safely navigate from its initial position to the goal (star) while avoiding collisions with other agents (red cars) that follow unknown control policies.
The blue car observes only the states of other agents. Its dynamics are discrete control-affine, given by:
\begin{align}
\label{eq:system-dynamics}
    \bfx_{t+1} = \bff(\bfx_t)+\bfg(\bfx_t) \bfa_t + \bfdelta(\bfx_t),
\end{align}
where state $\bfx \in \bbR^4$ contains position and velocity in $x$ and $y$ directions. Here, $\bff$ denotes the nominal drift dynamics, $\bfg$ is the input gain matrix, $\bfa$ is the control input (accelerations in $x$ and $y$ directions), and $\bfdelta$ represents unknown uncertainty in the dynamics.
The control policy first computes an optimal but potentially unsafe trajectory using an \ac{MPC} controller that ignores obstacles. Then, following~\cite[eq.~17]{cheng2020safe}, a \ac{CBF}-based controller is applied to compute the closest safe control input that respects the obstacle constraints. 
A quadratic program is formulated for the \ac{CBF}-based optimization problem, where the unknown uncertainty $\bfdelta$ is bounded inside a polytope. 
We use the worst-case error bounds predicted by DK2, KDK2, and KDK3 to define this polytope. Models use 5 hidden units.  
We evaluate performance over 50 trials, recording the number of \textbf{success} runs-reaching the goal without colliding with any other car, \textbf{collision}-crashes into another car, and cases where the agent becomes \textbf{stuck}-fails to reach the goal, either due to being blocked (e.g., deadlock) or overly conservative behavior resulting from excessive uncertainty estimation.
All models are trained offline for 300 epochs on 5000 noise-free data points collected. 
% Detailed results of this simulation are reported in Table~\ref{tbl:ControlExp}. 
We add uniform noise at three levels (5\%, 10\%, and 15\%) independently to each of the position, velocity, and acceleration, denoted by $\bfdelta_p$, $\bfdelta_v$, and $\bfdelta_a$. Considering all permutations of these noise levels results in 27 experimental setups, each repeated 50 times.
The dynamics in Equation~\ref{eq:system-dynamics} update position and velocity as $\bfp_{t+1} = \bfp_t + \bfv_t dt + \delta_p$ and $\bfv_{t+1} = \bfv_t - k_v |\bfv_t|^2 + k_d(|\bfv_t|+1)dt (\bfa + \delta_a) + \delta_v$, where $k_d$ is the acceleration scaling factor. 
The unknown part of the system is $\tilde \bfp_{t+1} = \delta_p$ and $\tilde \bfv_{t+1} = k_d(|\bfv_t|+1)dt \delta_a + \delta_v$.
The Lipschitz constant derivation for this unknown dynamics model with respect to velocity gives $\calL_{\partial \tilde p_{t+1}/\partial v} = 0$, and differentiating the velocity update gives $\calL_{\partial \tilde v_{t+1}/\partial v} = k_d \delta_a dt$. 
We use $k_d = 1.0$, and $dt = 0.1$ for the simulation step time.
The position and velocity disturbances, $\delta_p$ and $\delta_v$, are not captured by this Lipschitz constant; they are instead captured through the error-at-knots term (Sec.~\ref{sec:error-analysis}-3), since they enter the dynamics additively rather than through a term whose sensitivity the Lipschitz constant measures.
Detailed results of this simulation are reported in Table~\ref{tbl:ControlExp}. 

\begin{figure}
    \centering
    \includegraphics[width=0.98\linewidth]{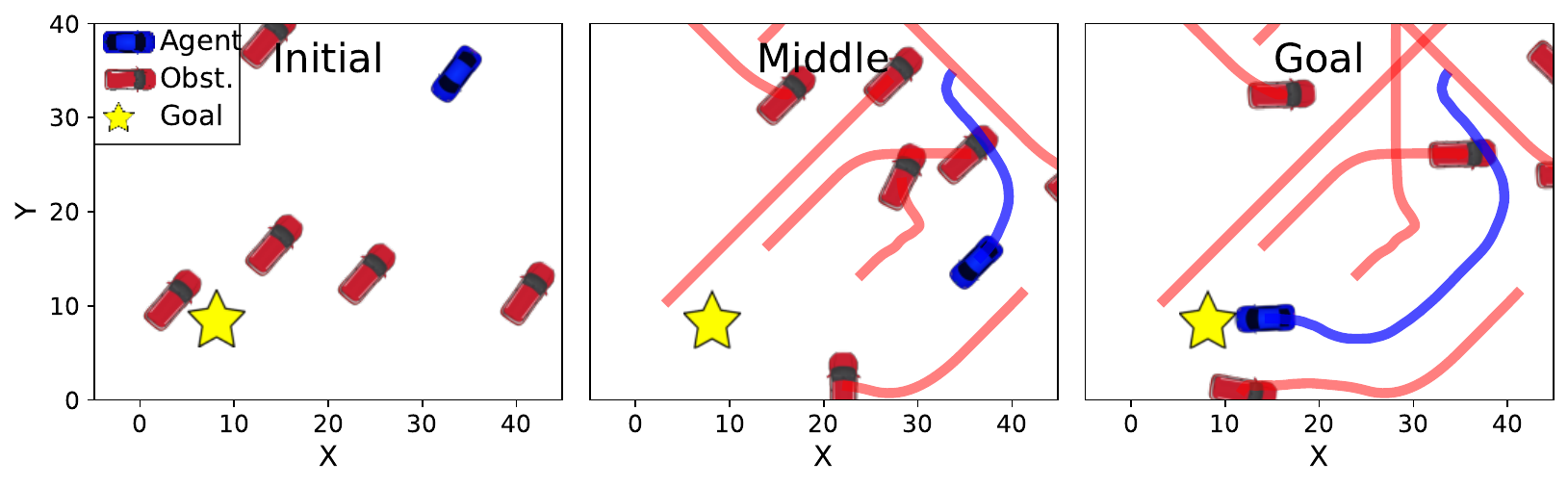}
    \caption{Three stages (Initial, Middle, Goal) of a successful trajectory using K-DAREK bounds, where the controlled agent (\blue{blue car}) safely navigates to the goal (star) while avoiding obstacle agents (\red{red cars}).}
    \label{fig:multiagent-traj}
\end{figure}

The \textit{Average} column reports the mean over all 27 setups corresponding to a specific value of $\bar{\bfdelta}_a$. 
The other three columns report the results for individual setups where $\bar{\bfdelta}_v = \bar{\bfdelta}_p \in$ (5\%, 10\%, and 15\%), with $\bar{\bfdelta}_a$ varying across rows. 
On average, \ac{K-DAREK} achieves a slightly higher success rate along with lower collision and stuck rates. A realization of safe navigation with \ac{K-DAREK} is illustrated in Fig.~\ref{fig:multiagent-traj}.

\begin{table*}
    \centering
    {\tiny
	\caption{
		Results for the multi-agent safe control experiment, where $\bar{\bfdelta}_a$, $\bar{\bfdelta}_v$, and $\bar{\bfdelta}_p$ are uniform disturbance rates on acceleration, velocity, and position. DK2 refers to two-layer DAREK, and KDK2 and KDK3 are two-layer and three-layer \ac{K-DAREK}, respectively, as described in the text. The KDK2 and KDK3 achieve a lower collision rate and slightly higher success rate than the DK2 model. 
		\label{tbl:ControlExp}}
	
        \setlength\extrarowheight{2pt}  
	\begin{tabular}{lcTTTVVVTTTVVV}
		\cline{1-14}
		\multicolumn{1}{c}{} & \multicolumn{1}{c}{} 
        &\multicolumn{3}{c}{$\bar{\bfdelta_v}=\bar{\bfdelta_p}=0.05 $}
        &\multicolumn{3}{c}{$\bar{\bfdelta_v}=\bar{\bfdelta_p}=0.10 $} 
        &\multicolumn{3}{c}{$\bar{\bfdelta_v}=\bar{\bfdelta_p}=0.15 $}
        &\multicolumn{3}{c}{Average} \\
		\cline{3-5}
        \cline{6-8}
        \cline{9-11}
        \cline{12-14}
            $\bar{\bfdelta_a}$ & Model & DK2 & KDK2 & KDK3
                          & DK2 & KDK2 & KDK3 
                          & DK2 & KDK2 & KDK3 
                          & DK2 & KDK2 & KDK3 \\
		\hline
		
        \multirow{3}{*}{$0.05 $}   
            &   Success   & 0.90 & 0.92 & 0.92 & 0.96 & 0.98 & 0.96 & 0.96 & 0.96 & 0.94 & 0.953 & 0.953 & 0.951 \\
		    &   Collision & 0.08 & 0.04 & 0.06 & 0.02 & 0.00 & 0.02 & 0.00 & 0.00 & 0.00 & 0.018 & 0.011 & 0.015 \\
            &   Stuck     & 0.02 & 0.04 & 0.02 & 0.02 & 0.02 & 0.02 & 0.04 & 0.04 & 0.06 & 0.028 & 0.035 & 0.033 \\
            \hline
        \multirow{3}{*}{$0.10 $}    
            &	Success   & 0.96 & 0.98 & 0.96 & 0.96 & 0.94 & 0.98 & 0.92 & 0.94 & 0.94 & 0.953 & 0.957 & 0.964 \\
            & Collision & 0.02 & 0.00 & 0.02 & 0.00 & 0.02 & 0.00 & 0.00 & 0.00 & 0.00 & 0.011 & 0.006 & 0.011 \\
            & Stuck     & 0.02 & 0.02 & 0.02 & 0.04 & 0.04 & 0.02 & 0.08 & 0.06 & 0.06 & 0.028 & 0.033 & 0.033 \\
            \hline
        \multirow{3}{*}{$0.15 $}   
            &	Success   & 0.92 & 0.94 & 0.98 & 0.96 & 0.96 & 0.98 & 0.96 & 0.96 & 0.92 & 0.947 & 0.955 & 0.955 \\
            &   Collision & 0.02 & 0.02 & 0.00 & 0.00 & 0.00 & 0.00 & 0.00 & 0.00 & 0.00 & 0.011 & 0.004 & 0.008 \\ 
            &   Stuck     & 0.06 & 0.04 & 0.02 & 0.04 & 0.04 & 0.02 & 0.04 & 0.04 & 0.08 & 0.031 & 0.035 & 0.033 \\
		\hline 
	\end{tabular}}
\end{table*}
\section{CONCLUSION}
We presented a general framework for quantifying distance-aware worst-case error in a hybrid model that combines deep dense \acp{NN} and spline-based networks, namely \ac{K-DAREK}.
We derived an explicit error bound for the \ac{MLP} block using its Lipschitz constant, thereby linking test point uncertainty to proximity to the training data. For the spline block, we redefined the DAREK formulation to align with this purpose based on the selection of knots derived from the training set and the accuracy of predicted values at those knots. By combining these components, we obtained an overall model error bound that captures both local interpolation uncertainty and propagated transformation error through the model. 
This framework yields an error bound that is distance-aware and does not require sampling, ensembling, or distributional assumptions about the data, and is based on Lipschitz continuity.
We further evaluated whether the component-wise construction of K-DAREK's bound yields joint distance-awareness in higher dimensions through the SDA metric on the face-detection task, where K-DAREK reaches SDA values of approximately 95\%. Also, on real data regression tasks, K-DAREK achieves zero {or near-zero} coverage violations {across most datasets,} while achieving competitive accuracy compared to baselines.
Our model avoids overgeneralization {(feature collapse) observed in SNGP,}
and its behavior is consistent in low and high amounts of data. Additionally, in multi-agent safe control tasks, K-DAREK improves success rates and reduces collisions.
{We also conduct an ablation study to separately quantify the effects of the MLP block, spectral normalization, and knot-selection strategy, and spline capacity on accuracy, empirical coverage, and average bound width.
We additionally quantify how violation rate degrades as the assumed Lipschitz constant is underestimated, showing that the guarantee's reliability depends directly on the accuracy of this estimate, consistent with the conditional nature of the guarantee discussed in Sec.~\ref{sec:method}.}

This framework provides a theoretically grounded understanding of the reliability of prediction and error estimation, as well as computational efficiency, paving the way for robust learning in settings that demand 
deterministic, worst-case error bounds with an assumed Lipschitz constant or risk-aware predictions.
However, several limitations of the current framework suggest directions for future work. First, the analysis assumes that the target function is continuous and Lipschitz, and it does not directly apply to predictors with discontinuous or Boolean outputs. Second, spectral normalization controls the Lipschitz constant of each MLP block, but the spline block is not explicitly constrained in the same way. 
Moreover, the current formulation divides the Lipschitz budget uniformly among splines within a group (Remark~\ref{rem:lip-div}). 
A non-uniform allocation that accounts for each spline's local curvature could yield a tighter overall bound. 
Third, knots are selected from the training data, which makes the hybrid network data-adaptive and ties the resulting error bound to observed samples. 
We compare several alternative knot-selection strategies in Table~\ref{tbl:ablation}; a more exhaustive comparison across additional tasks is left for future work (Remark~\ref{rem:knot-sel}).
Fourth, the theoretical guarantee relies on knowledge of the target function's Lipschitz constant. 
In practice, this constant may only be approximated (as in our simulations); 
{our sensitivity analysis (Sec.~\ref{sec:experiment}) quantifies how the violation rate degrades under such approximation, and accounting for this approximation more formally in the error bound is left for future work.}
Fifth, restricting the Chebyshev expansion of the MLP block in \ac{KKAN} architecture to its first two degrees simplifies the error analysis at the cost of expressiveness; a higher-order expansion with a corresponding error analysis is left for future work. Finally, the present work focuses on worst-case analysis; developing hybrid worst-case/probabilistic methods to tighten the bounds in data-rich scenarios is left for future research.

\bibliographystyle{IEEEtran}
\bibliography{bib/main}

\end{document}